\documentclass{article}

\PassOptionsToPackage{numbers}{natbib}

\usepackage[final]{neurips_2025}
\usepackage{wrapfig}
\usepackage{adjustbox}
\usepackage[utf8]{inputenc} 
\usepackage[T1]{fontenc}    
\usepackage{hyperref}       
\usepackage{url}            
\usepackage{booktabs}       
\usepackage{amsfonts}       
\usepackage{nicefrac}       
\usepackage{microtype}      
\usepackage{xcolor}   
\usepackage{amsmath,amssymb,amsthm,amsfonts}
\usepackage{dsfont}
\usepackage{wrapfig}

\usepackage{booktabs}
\usepackage{multirow}
\usepackage{multicol}
\usepackage{graphicx}
\usepackage{caption}
\usepackage{subcaption}
\usepackage{algorithm,algorithmic}
\usepackage{autobreak}
\allowdisplaybreaks
\newtheorem{theorem}{Theorem}

\title{Explainable Reinforcement Learning from Human Feedback to Improve Alignment}

\author{%
Shicheng Liu$^{1}$, Siyuan Xu$^{1}$, Wenjie Qiu$^{2}$, Hangfan Zhang$^{3}$, Minghui Zhu$^{1}$ \\
$^{1}$Department of Electrical Engineering, Pennsylvania State University \\
$^{2}$Department of Computer Science, Rutgers University \\
$^{3}$College of Information Sciences and Technology, Pennsylvania State University \\
\texttt{\{sfl5539, spx5032, hbz5148, muz16\}@psu.edu}, \texttt{wenjie.qiu@rutgers.edu}
}

\begin{document}

\maketitle

\begin{abstract}
  A common and effective strategy for humans to improve an unsatisfactory outcome in daily life is to find a cause of this outcome and correct the cause. In this paper, we investigate whether this human improvement strategy can be applied to improving reinforcement learning from human feedback (RLHF) for alignment of language models (LMs). In particular, it is observed in the literature that LMs tuned by RLHF can still output unsatisfactory responses. This paper proposes a method to improve the unsatisfactory responses by correcting their causes. Our method has two parts. The first part proposes a post-hoc explanation method to explain why an unsatisfactory response is generated to a prompt by identifying the training data that lead to this response. We formulate this problem as a constrained combinatorial optimization problem where the objective is to find a set of training data closest to this prompt-response pair in a feature representation space, and the constraint is that the prompt-response pair can be decomposed as a convex combination of this set of training data in the feature space. We propose an efficient iterative data selection algorithm to solve this problem. The second part proposes an unlearning method that improves unsatisfactory responses to some prompts by unlearning the training data that lead to these unsatisfactory responses and, meanwhile, does not significantly degrade satisfactory responses to other prompts. Experimental results demonstrate that our algorithm can improve RLHF.
\end{abstract}

\section{Introduction}

Reinforcement learning from human feedback (RLHF) \cite{casper2023open,kaufmann2023survey,christiano2017deep} is a predominant approach to align language models (LMs) with human preference. It learns a reward model from human preference data and uses the reward model to tune an LM \cite{ziegler2019fine,ouyang2022training,bai2022training}. The tuned LM can be evaluated on a set of validation prompts $\{\bar{x}^{(i)}\}_{i=1}^M$ to generate responses $\{\bar{y}^{(i)}\}_{i=1}^M$, thereby forming the validation data $\bar{\mathcal{D}}=\{(\bar{x}^{(i)},\bar{y}^{(i)})\}_{i=1}^M$. It is expected that all responses in $\bar{\mathcal{D}}$ should align with human preference. However, it is observed in the literature that LMs tuned by RLHF can still generate unsatisfactory responses, such as harmful \cite{daisafe,yuangpt}, inaccurate \cite{chen2024accuracy,leerlaif} and redundant \cite{chen2024not,chiang2024over} responses. We also include an experiment (in Appendix \ref{subsec: unsatisfactory labels}) to validate that RLHF can still generate unsatisfactory responses. Therefore, we can partition the validation data $\bar{\mathcal{D}}$ into an unsatisfactory subset $\bar{\mathcal{D}}_{\text{u}}$ (with unsatisfactory responses and associated validation prompts) and a satisfactory subset $\bar{\mathcal{D}}\setminus\bar{\mathcal{D}}_{\text{u}}$. These unsatisfactory responses in $\bar{\mathcal{D}}_{\text{u}}$ can be due to various reasons, such as missing or misleading information in the human preference data. In this paper, we study the case where the preference data includes misleading information. This setting is common in practice. One example is that the preference data is usually collected from diverse sources and thus some preference data can be useful for some prompts but misleading to other prompts \cite{casper2023open,pitis2023failure}. Another example is that the preference data can be noisy, where the labels of some actually preferred and dispreferred responses are flipped due to the bias and errors of human annotators or malicious noise injection \cite{kong2024perplexity,cheng2024rime}. 
\begin{figure}[t]
    \centering
    \includegraphics[width=0.99\linewidth]{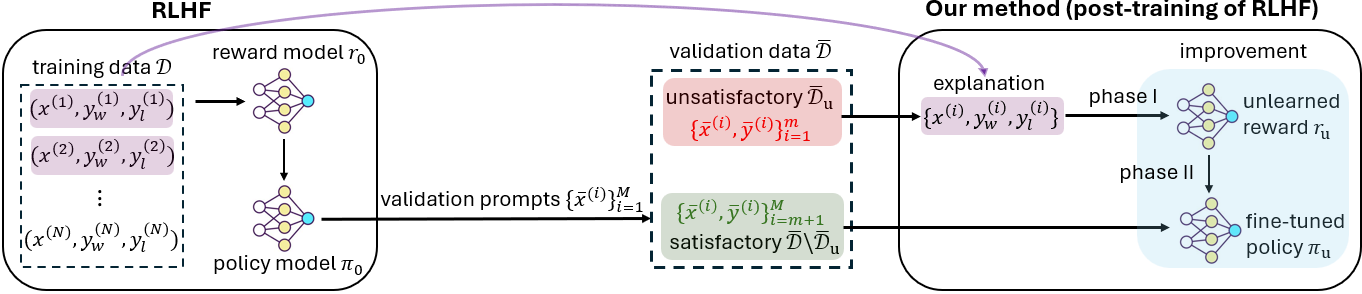}
    \caption{An illustration of our method. Our method can be viewed as a post-training of RLHF. RLHF learns a reward model $r_0$ and a policy model $\pi_0$. The policy $\pi_0$ is evaluated on $\{\bar{x}^{(i)}\}_{i=1}^M$ to form the validation data $\bar{\mathcal{D}}$ that consists of an unsatisfactory subset $\bar{\mathcal{D}}_{\text{u}}$ and a satisfactory subset $\bar{\mathcal{D}}\setminus\bar{\mathcal{D}}_{\text{u}}$. Our method has two parts. The first part explains why $\pi_0$ generates the unsatisfactory responses to the prompts in $\bar{\mathcal{D}}_{\text{u}}$ by identifying the training data that lead $\pi_0$ to generate these responses. The second part includes two phases to improve $\pi_0$. Phase one unlearns the identified training data from the original reward model $r_0$ to obtain an unlearned reward model $r_{\text{u}}$. Phase two fine-tunes $\pi_0$ to maximize the unlearned reward model $r_{\text{u}}$ for the validation prompts in $\bar{\mathcal{D}}_{\text{u}}$, while staying close to $\pi_0$ for the validation prompts in $\bar{\mathcal{D}}\setminus\bar{\mathcal{D}}_{\text{u}}$.}
    \label{fig: post training strucutre}
\end{figure}

This paper proposes a method that further improves the LMs tuned by RLHF to better align with human preference. Our method is inspired by an effective strategy that humans commonly use in daily life to improve an unsatisfactory outcome, where humans first find a cause of the outcome and then correct the cause to improve the outcome. Our method has two parts. The first part proposes a post-hoc explanation method to explain why the LM generates the unsatisfactory responses in $\bar{\mathcal{D}}_{\text{u}}$ by identifying the training data that lead to these responses. The second part improves the unsatisfactory responses by unlearning the identified training data in the first part, and does not significantly degrade satisfactory responses in $\bar{\mathcal{D}}\setminus\bar{\mathcal{D}}_{\text{u}}$. Our method (illustrated in Figure \ref{fig: post training strucutre}) can be viewed as a post-training of RLHF as our method operates after RLHF has tuned an LM.

The first part proposes a post-hoc explanation method to identify a set of training data that leads the LM (tuned by RLHF) to generate the unsatisfactory response $\bar{y}$ to the prompt $\bar{x}$ in $\bar{\mathcal{D}}_{\text{u}}$. We first mathematically derive an insight that if $(\bar{x},\bar{y})$ is either close to, or can be expressed as a convex combination of, some training data in a feature space, then the response $\bar{y}$ is likely to be generated to $\bar{x}$. Using this insight as a guideline, we formulate a constrained combinatorial optimization problem where the objective is to find a subset of training data whose features are closest to the feature of $(\bar{x},\bar{y})$, and the constraint is that the feature of $(\bar{x},\bar{y})$ can be decomposed as a convex combination of the features of this set of training data. This optimization problem is NP-hard and brute-force enumeration of all possible subsets of training data is computationally intractable, as it results in an exponential computational complexity with respect to the number of training data \cite{garey1983computers}. To address this issue, we propose an efficient iterative data selection algorithm which starts with the closest training data and keeps adding data to expand the set until the constraint is satisfied. We prove that the computational complexity of the proposed algorithm is polynomial.

The second part uses the subset of identified training data in the first part to improve the unsatisfactory responses in $\bar{\mathcal{D}}_{\text{u}}$ and, meanwhile, does not significantly degrade the satisfactory responses in $\bar{\mathcal{D}}\setminus\bar{\mathcal{D}}_{\text{u}}$. Specifically, the second part includes two phases: reward unlearning and policy fine-tuning. The reward unlearning phase unlearns the identified training data from the reward model of RLHF to obtain an unlearned reward model by minimizing the log-likelihood of this identified subset of training data. The policy fine-tuning phase fine-tunes the policy of RLHF to maximize the unlearned reward for the validation prompts in $\bar{\mathcal{D}}_{\text{u}}$ and restricts the KL divergence from the policy of RLHF for the validation prompts in $\bar{\mathcal{D}}\setminus\bar{\mathcal{D}}_{\text{u}}$.

\textbf{Contribution statement}. This paper proposes a method, applied after RLHF, that leverages post-hoc explanations to further improve alignment. Our contributions are twofold.

First, we study the explanation problem of identifying the training data that lead to unsatisfactory responses. The major novelty is that we mathematically derive an insight into what kind of unsatisfactory responses is likely to be output by the LM and, based on this insight, we mathematically formalize this problem as a constrained combinatorial optimization problem. While the formulated optimization problem is NP-hard, we propose an efficient algorithm with polynomial computational complexity to solve this problem.

Second, we improve the unsatisfactory responses by unlearning the identified training data and do not significantly degrade other satisfactory responses. We propose a novel unlearning method that first unlearns the reward model and then fine-tunes the policy model. We conduct extensive experiments to demonstrate that our proposed method can further improve the LMs tuned by RLHF.

\section{Related works}
Due to the space limit, here we only discuss related works on post-hoc explainability and language model unlearning. We include discussions on more related works in Appendix \ref{sec: more related works}.

\textbf{Post-hoc explainability (PHE)}. PHE provides explanations for why a \emph{trained} model outputs a certain decision for an input. Existing approaches can be broadly categorized into feature-based and example-based methods. Feature-based methods \cite{datta2016algorithmic,ribeiro2016should,fong2017interpretable} learn an importance score for each input feature and use this score to reflect the importance of each feature for the trained model to output its decision. Example-based methods \cite{crabbe2021explaining,caruana1999case,sun2024accountability} find relevant examples of the input and use the decision of the relevant examples to explain the decision of the input. This category is inspired by the observation that humans usually use relevant experience to interpret a new thing \cite{crabbe2021explaining}. Our explanation is an example-based explanation. However, the aforementioned example-based methods cannot be applied to our problem because they require the relevant examples to have same data structure as the data to be explained. In our case, the data to be explained is a prompt-response pair but the training data we use to explain includes a prompt and a response comparison.

\textbf{Language model unlearning (LMU)}. LMU \cite{yao2023large,yao2024machine,liu2024towards} is proposed as an alternative to RLHF for tuning LMs by removing negative or harmful information from them. Specifically, LMU assumes access to a pre-collected set of negative examples and aims to directly unlearn these examples from the LMs. However, these unlearning methods cannot be applied to our setting because we cannot directly unlearn the training data from the policy/language model. The reason is that training data consists of a prompt and a response comparison, while the policy/language model outputs a response (instead of a response comparison) given a prompt. This mismatch prohibits direct unlearning from the policy model. Therefore, we propose a novel unlearning method that first unlearns the reward model and then fine-tunes the policy model under the unlearned reward model.

\textbf{Approaches to improve alignment}. There are other approaches proposed to improve alignment of RLHF. Specifically, the paper \cite{shen2024improving} proposes to train the policy under a contrastive reward function (subtracting the learned reward model by the reward model of the SFT policy) to improve alignment. The paper \cite{zhang2024improving} proposes to learn ensemble reward functions (a combination of multiple reward functions) to improve reward learning accuracy and thus improve alignment. The paper \cite{chegini2024model} proposes to average two independent SFT policies as the reference policy to allow for larger deviation from the SFT policies to improve alignment.

\section{Preliminaries}\label{sec: preliminaries}
RLHF typically includes three stages: 1) \textbf{supervised fine-tuning (SFT)}, where high-quality demonstration data is used to fine-tune a pre-trained model in a supervised manner to get a model $\pi_{\text{SFT}}$; 2) \textbf{reward modeling (RM)}, where preference data is used to train a reward model; 3) \textbf{Reinforcement learning (RL)}, where the SFT model $\pi_{\text{SFT}}$ is further fine-tuned by running RL to optimize the reward model. In this paper, we assume that $\pi_{\text{SFT}}$ is given, and we elaborate the last two stages.

\textbf{Reward modeling (RM)}. Given a prompt $x$, the SFT model $\pi_{\text{SFT}}$ can generate a pair of responses $(y_1,y_2)$. Human annotators are instructed to choose the response they prefer from the pair $(y_1,y_2)$, resulting in $y_w\succ y_l$ where $y_w$ and $y_l$ are respectively the preferred and dispreferred responses. We assume the access to a set of preference data $\mathcal{D}=\{x^{(i)},y_w^{(i)},y_l^{(i)}\}_{i=1}^N$, and train a reward model $r_{\theta}$ by maximizing the log-likelihood of the preference $\mathcal{D}$ under Bradley-Terry model \cite{bradley1952rank}:
\begin{equation}
    \max_{\theta}\; L(\theta,\mathcal{D})\triangleq \frac{1}{N}\sum_{i=1}^N\Big[\log\sigma(r_{\theta}(x^{(i)},y_w^{(i)})-r_{\theta}(x^{(i)},y_l^{(i)}))\Big], \label{eq: reward model learning}
\end{equation}
where $\sigma$ is the sigmoid function, and $P(y_w\succ y_l)=\sigma(r_{\theta}(x,y_w)-r_{\theta}(x,y_l))$ is the Bradley-Terry model. We denote the reward model learned in this stage by $r_{\theta_0}$.

\textbf{Reinforcement learning (RL)}. With the learned reward model $r_{\theta_0}$, we aim to further fine-tune the SFT model $\pi_{\text{SFT}}$ to align with human preference by solving the following RL problem:
\begin{equation*}
    \max_{\pi}\; E_{x\sim\mathcal{D},y\sim \pi(\cdot|x)}[r_{\theta_0}(x,y)]-\beta E_{x\sim\mathcal{D}}[D_{\text{KL}}(\pi(\cdot|x)||\pi_{\text{SFT}}(\cdot|x))],
\end{equation*}
where $\beta$ is a hyper-parameter controlling the deviation of the learned policy $\pi$ from the SFT model $\pi_{\text{SFT}}$, and $D_{\text{KL}}(\pi(\cdot|x)||\pi_{\text{SFT}}(\cdot|x))\triangleq E_{y\sim \pi(\cdot|x)}[\log\pi(y|x)-\log\pi_{\text{SFT}}(y|x)]$ is the KL divergence between the policy $\pi$ and SFT model $\pi_{\text{SFT}}$. We denote the learned policy model by $\pi_0$.

We can evaluate the responses $\bar{y}$ of $\pi_0$ to the validation prompts $\bar{x}$. We use $\bar{\mathcal{D}}=\{\bar{x}^{(i)},\bar{y}^{(i)}\}_{i=1}^M$ to denote the set of validation prompts and their corresponding responses generated by $\pi_0$. It is expected that all the responses generated by $\pi_0$ are satisfactory, however, it is observed in the literature and our experiment (in Appendix \ref{subsec: unsatisfactory labels}) that $\pi_0$ can generate unsatisfactory responses. In our case, we use open-source reward models to score responses and identify responses with scores below a threshold as unsatisfactory. The threshold is picked by human evaluations (detailed in \ref{subsec: unsatisfactory labels}). Therefore, we partition the set $\bar{\mathcal{D}}$ into two subsets. The unsatisfactory subset $\bar{\mathcal{D}}_{\text{u}}=\{\bar{x}^{(i)},\bar{y}^{(i)}\}_{i=1}^m$ includes the unsatisfactory responses in $\bar{\mathcal{D}}$ and their associated validation prompts. The rest $\bar{\mathcal{D}}\setminus\bar{\mathcal{D}}_{\text{u}}$ includes the satisfactory responses in $\bar{\mathcal{D}}$ and their associated validation prompts.

\section{Explainable reinforcement learning from human feedback}\label{sec: explanation}
\begin{wrapfigure}[11]{r}{0.6\textwidth}
    \vspace{-4mm}
    \centering
    \includegraphics[width=0.59\textwidth]{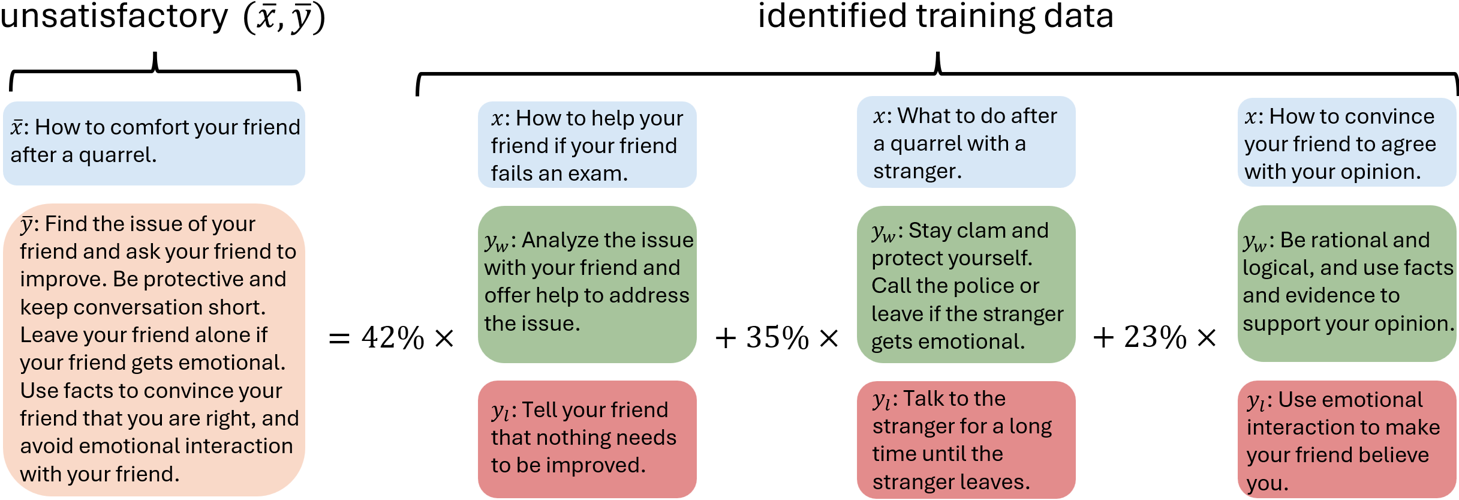}
    \caption{An example of explaining an unsatisfactory response $\bar{y}$ to a prompt $\bar{x}$ using a set of three training data.}
\end{wrapfigure}
This section proposes an example-based post-hoc explanation method. Inspired by the way humans use relevant experience to interpret new things, example-based methods use relevant examples to explain why a learned model generates a certain output \cite{crabbe2021explaining,caruana1999case,sun2024accountability}. In this section, we explain why the RLHF policy model $\pi_0$ generates an unsatisfactory response $\bar{y}$ to a prompt $\bar{x}$ by identifying the training data in $\mathcal{D}$ that lead to this response. Given that the policy $\pi_0$ generates the response $\bar{y}$ to the prompt $\bar{x}$, it means that the learned reward model $r_{\theta_0}$ assigns a high reward $r_{\theta_0}(\bar{x},\bar{y})$ to the prompt-response pair $(\bar{x},\bar{y})$. Therefore, it suffices to explain why $(\bar{x},\bar{y})$ has a high reward. To facilitate understanding, we first consider the linear reward case $r_{\theta_0}(x,y)=\theta_0^{\top}\phi(x,y)$ to illustrate our method, where $\theta_0$ is the reward parameter and $\phi(\cdot,\cdot)$ is a feature function that maps the prompt-response pair to a feature vector. \emph{We emphasize that our explanation method is not limited to the linear reward case, and we will later discuss how to directly use our method in general reward cases}.

We aim to identify the training data that contribute to the high reward $r_{\theta_0}(\bar{x},\bar{y})$: the identified examples contribute to the high value of $r_{\theta_0}(\bar{x},\bar{y})$, and since RL optimizes the LM to maximize reward, a high reward increases the likelihood that $\bar{y}$ is generated. Therefore, the identified examples (indirectly) contribute to the generation of $\bar{y}$. For this purpose, we first reason about what kind of prompt-response pairs has high rewards. Recall from the reward learning problem \eqref{eq: reward model learning} where we aim to optimize $\theta$ to maximize the log-likelihood over the training set $\mathcal{D}$. We can reformulate the log-likelihood function in \eqref{eq: reward model learning} as (derived in Appendix \ref{appendixsec: derivation of reformulation}):
\begin{equation*}
    L(\theta,\mathcal{D})=\frac{1}{N}\sum_{i=1}^N\Big[\theta^{\top}\Big(\phi(x^{(i)},y_w^{(i)})-\phi(x^{(i)},y_l^{(i)})\Big)-\log\Big(e^{\theta^{\top}(\phi(x^{(i)},y_w^{(i)})-\phi(x^{(i)},y_l^{(i)}))}+1\Big)\Big].
\end{equation*}
Since the above log-likelihood function $L(\theta,\mathcal{D})$ is strictly monotonically increasing in the inner product $\theta^{\top}(\phi(x^{(i)},y_w^{(i)})-\phi(x^{(i)},y_l^{(i)}))$ (proved in Appendix \ref{appendixsec: derivation of reformulation}), maximizing the log-likelihood is equivalent to maximizing the inner product over the preference training set $\mathcal{D}$. Therefore, the feature comparisons $\{\phi(x^{(i)},y_w^{(i)})-\phi(x^{(i)},y_l^{(i)})\}_{i=1}^N$ of the training data have high rewards, i.e., $\theta_0^{\top}(\phi(x^{(i)},y_w^{(i)})-\phi(x^{(i)},y_l^{(i)}))$ is high. If a prompt-response pair $(\bar{x},\bar{y})$ has a feature $\phi(\bar{x},\bar{y})$ closely aligned with the feature comparisons of some training data, this prompt-response pair is expected to have a high reward.

The above insight motivates two distinct intuitions for finding training data to explain why $(\bar{x},\bar{y})$ has a high reward:

\textbf{Intuition I (nearest neighbors).} We find $n$ ($1\leq n\leq N$) training data that is closest to $(\bar{x},\bar{y})$ in the feature space, i.e., minimum Euclidean distance $||\phi(x^{(i)},y_w^{(i)})-\phi(x^{(i)},y_l^{(i)})-\phi(\bar{x},\bar{y})||$. Since the reward function is continuous in $\phi$, the feature $\phi(\bar{x},\bar{y})$ with small Euclidean distance from the feature comparisons of some training data should also have a high reward. 

\textbf{Intuition II (decomposition).} We find $n$ training data to decompose $(\bar{x},\bar{y})$ in the feature space such that $\phi(\bar{x},\bar{y})=\sum_{i=1}^n \omega^{(i)}(\phi(x^{(i)},y_w^{(i)})-\phi(x^{(i)},y_l^{(i)}))$ where $\omega^{(i)}\in[0,1]$ and $\sum_{i=1}^{n}\omega^{(i)}=1$. Since the feature comparisons of the $n$ training data have high rewards, their convex combination $\phi(\bar{x},\bar{y})$ should also have a high reward.

Intuition I is easy to compute but can be imprecise. If the nearest neighbor is still relatively far from $(\bar{x},\bar{y})$ (i.e., $||\phi(x^{(i)},y_w^{(i)})-\phi(x^{(i)},y_l^{(i)})-\phi(\bar{x},\bar{y})||$ is large), the reward difference between the neighbor and $(\bar{x},\bar{y})$ can also be large, leading to an inaccurate explanation.

Intuition II can precisely explain the reward value $r_{\theta_0}(\bar{x},\bar{y})$ even if the neighbors are far, because it matches the feature $\phi(\bar{x},\bar{y})$, ensuring that $r_{\theta_0}(\bar{x},\bar{y})=\sum_{i=1}^n \omega^{(i)}(r_{\theta_0}(x^{(i)},y_w^{(i)})-r_{\theta_0}(x^{(i)},y_l^{(i)}))$. However, finding a decomposition can be an ill-posed problem because there could be multiple feasible convex combinations that can match $\phi(\bar{x},\bar{y})$.

We propose a method that combines the strengths of both intuitions by learning a decomposition of $\phi(\bar{x},\bar{y})$ that is closest to $\phi(\bar{x},\bar{y})$. However, such a decomposition is infeasible if $\phi(\bar{x},\bar{y})$ does not lie within the convex hull of the feature comparisons from the training set. Therefore, we first project the feature vector $\phi(\bar{x},\bar{y})$ onto the convex hull. Specifically, the convex hull of the training data feature comparisons is defined as $\mathcal{C}_{\phi}(\mathcal{D})\triangleq \{\sum_{i=1}^{|\mathcal{D}|} \omega^{(i)}(\phi(x^{(i)},y_w^{(i)})-\phi(x^{(i)},y_l^{(i)}))|\omega^{(i)}\in[0,1],\sum_{i=1}^{|\mathcal{D}|}\omega^{(i)}=1\}$ where $|\mathcal{D}|=N$ is the cardinality of the training set $\mathcal{D}$. We project $\phi(\bar{x},\bar{y})$ onto $\mathcal{C}_{\phi}(\mathcal{D})$ to obtain the projected feature vector $\hat{\phi}(\bar{x},\bar{y})$ by solving the quadratic program:
\begin{equation}
    \hat{\phi}(\bar{x},\bar{y})=\mathop{\arg\min}_{v\in\mathcal{C}_{\phi}(\mathcal{D})}||v-\phi(\bar{x},\bar{y})||^2, \label{eq: projection qudratic program}
\end{equation}
where the projection $\hat{\phi}(\bar{x},\bar{y})=\phi(\bar{x},\bar{y})$ if $\phi(\bar{x},\bar{y})\in\mathcal{C}_{\phi}(\mathcal{D})$. Next, we aim to decompose the projected feature vector $\hat{\phi}(\bar{x},\bar{y})$ as a convex combination of feature comparisons from a subset of training data. Among all feasible convex combinations, we find the one that is closest to $\hat{\phi}(\bar{x},\bar{y})$. Formally, we propose to solve the following constrained combinatorial optimization problem:
\begin{align}
    &\min_{\mathcal{S}\subseteq \mathcal{D}}\;\sum_{i=1}^{|\mathcal{S}|}||\phi(x^{(i)},y_w^{(i)})-\phi(x^{(i)},y_l^{(i)})-\hat{\phi}(\bar{x},\bar{y})||,\notag\\
    &\text{s.t.} \quad \hat{\phi}(\bar{x},\bar{y})\in \mathcal{C}_{\phi}(\mathcal{S}), \quad (x^{(i)},y_w^{(i)},y_l^{(i)})\in\mathcal{S},\; i=1,\cdots,|\mathcal{S}|. \label{eq: combinatorial optimization}
\end{align}
The problem \eqref{eq: combinatorial optimization} aims to find a subset $\mathcal{S}$ of the training set $\mathcal{D}$ such that $\hat{\phi}(\bar{x},\bar{y})$ lies within the convex hull $\mathcal{C}_{\phi}(\mathcal{S})$ and the sum of Euclidean distances between $\hat{\phi}(\bar{x},\bar{y})$ and the feature comparison of each training data in $\mathcal{S}$ is minimized.

\textbf{Remark on general reward cases}. Although we derive the problem \eqref{eq: combinatorial optimization} assuming a linear reward model for clarity, the problem \eqref{eq: combinatorial optimization} can be directly applied to general reward cases. In practice, reward models are usually initialized from a SFT model $\pi_{\text{SFT}}$ with an additional linear layer on top of the final transformer layer \cite{ziegler2019fine,rafailov2024direct}. After the reward model is trained, given a prompt-response pair $(x,y)$, the output of the final transformer layer can be used as the feature representation $\phi(x,y)$. In this case, the reward of $(x,y)$ is also linear of $\phi(x,y)$. If the reward model is a neural network where the last layer is a fully connected (linear) layer, we can use the output of the second-to-last layer as the feature representation.

\subsection{Minimum-distance-based iterative training data selection}
The problem \eqref{eq: combinatorial optimization} is guaranteed to have an optimal solution because 1) its feasible set is non-empty as $\hat{\phi}(\bar{x},\bar{y})\in \mathcal{C}_{\phi}(\mathcal{D})$; 2) there are finitely many choices of $\mathcal{S}$ since $\mathcal{D}$ is a finite set. However, the space of all possible choices of $\mathcal{S}$ can be extremely large because it grows exponentially with the cardinality of $\mathcal{D}$; therefore, it is intractable to search over the entire space to find the optimal set $\mathcal{S}$. To address this issue, we propose an iterative training data selection algorithm where we start from an empty set $\mathcal{S}$ and iteratively add new training data whose feature comparisons are close to $\hat{\phi}(\bar{x},\bar{y})$ until $\mathcal{C}_{\phi}(\mathcal{S})$ contains $\hat{\phi}(\bar{x},\bar{y})$. This algorithm is efficient because it avoids searching over the entire space of all possible choices of $\mathcal{S}$. Instead, it only searches over a much smaller local space around $\hat{\phi}(\bar{x},\bar{y})$. We elaborate the algorithm as follows:
\begin{algorithm}[t]
\caption{Explainable reinforcement learning from human feedback (XRLHF)}
\textbf{Input}: A prompt-response pair $(\bar{x},\bar{y})$ to be explained, the training set $\mathcal{D}$, and an empty set $\mathcal{S}$.\\
\textbf{Output}: The set $\mathcal{S}$ of training data that leads to the response $\bar{y}$ to the prompt $\bar{x}$ and the corresponding decomposition coefficients $\{\omega^{(i)}\}_{i=1}^{|\mathcal{S}|}$.

\begin{algorithmic}[1]\label{alg: XRLHF}
\STATE Compute the projected feature vector $\hat{\phi}(\bar{x},\bar{y})$ by solving the quadratic program \eqref{eq: projection qudratic program}. \label{line: find projected feature vector}
\STATE Rank the training data $(x,y_w,y_l)$ in $\mathcal{D}$ by the distance $||\phi(x,y_w)-\phi(x,y_l)-\hat{\phi}(\bar{x},\bar{y})||$. \label{line: rank by distance}
\WHILE{$\hat{\phi}(\bar{x},\bar{y})\notin \mathcal{C}_{\phi}(\mathcal{S})$}
\STATE Pick the nearest training data $(x,y_w,y_l)$ in $\mathcal{D}\setminus\mathcal{S}$ and add this data point $\mathcal{S}=\mathcal{S}\cup (x,y_w,y_l)$. \label{line: nearest training data and add}
\STATE Check the condition $\hat{\phi}(\bar{x},\bar{y})\in \mathcal{C}_{\phi}(\mathcal{S})$ and find the corresponding decomposition coefficients $\{\omega^{(i)}\}_{i=1}^{|\mathcal{S}|}$ if condition satisfied, by solving the linear program \eqref{eq: linear program}.
\ENDWHILE
\end{algorithmic}
\end{algorithm}

Given a prompt-response pair $(\bar{x},\bar{y})$ that requires explanation, Algorithm \ref{alg: XRLHF} first computes its projected feature vector $\hat{\phi}(\bar{x},\bar{y})$ by solving the quadratic program \eqref{eq: projection qudratic program} in line \ref{line: find projected feature vector} and ranks all the training data in $\mathcal{D}$ by the Euclidean distance $||\phi(x,y_w)-\phi(x,y_l)-\hat{\phi}(\bar{x},\bar{y})||$ (line \ref{line: rank by distance}). Then Algorithm \ref{alg: XRLHF} solves the problem \eqref{eq: combinatorial optimization} by iteratively picking the nearest training data. At each iteration, Algorithm \ref{alg: XRLHF} finds the closest training data in $\mathcal{D}\setminus\mathcal{S}$ according to the ranking and adds this training data to $\mathcal{S}$ (line \ref{line: nearest training data and add}). The algorithm checks whether the convex hull $\mathcal{C}_{\phi}(\mathcal{S})$ corresponding to the current set $\mathcal{S}$ already contains $\hat{\phi}(\bar{x},\bar{y})$ by solving the following linear program:
\begin{equation}\label{eq: linear program}
    \min 0,\quad \text{s.t.} \;\;\hat{\phi}(\bar{x},\bar{y})=\sum_{i=1}^{|\mathcal{S}|} \omega^{(i)}(\phi(x^{(i)},y_w^{(i)})-\phi(x^{(i)},y_l^{(i)})),\;\;\omega^{(i)}\in[0,1],\;\;\sum_{i=1}^{|\mathcal{S}|}\omega^{(i)}=1.
\end{equation}
The problem \eqref{eq: linear program} is a linear programming feasibility problem that not only checks whether $\hat{\phi}(\bar{x},\bar{y})\in \mathcal{C}_{\phi}(\mathcal{S})$ but also provides the decomposition coefficients $\{\omega^{(i)}\}_{i=1}^{|\mathcal{S}|}$ if a decomposition is feasible. If multiple decompositions are feasible in $\mathcal{S}$, we choose the one closest to $\hat{\phi}(\bar{x},\bar{y})$.

\begin{theorem}\label{thm: computational complexity of explanation}
    The computational complexity of Algorithm \ref{alg: XRLHF} is $O(N\log N+Nd+N^2d+N^{3.5}+N^{4.5})$ where $N$ is the number of training data in $\mathcal{D}$ and $d$ is the feature dimension.
\end{theorem}

\section{The improvement via unlearning}\label{sec: unlearning}
This section improves the alignment of $\pi_0$ by leveraging the explanation method in Section \ref{sec: explanation}. Recall from Section \ref{sec: preliminaries} that the validation data $\bar{\mathcal{D}}=\{\bar{x}^{(i)},
\bar{y}^{(i)}\}_{i=1}^M$ is partitioned into an unsatisfactory subset $\bar{\mathcal{D}}_{\text{u}}=\{\bar{x}^{(i)},\bar{y}^{(i)}\}_{i=1}^m$ and a satisfactory subset $\bar{\mathcal{D}}\setminus\bar{\mathcal{D}}_{\text{u}}$. We aim to improve $\pi_0$ so that its responses to the prompts in $\bar{\mathcal{D}}_{\text{u}}$ are improved, while its responses to the prompts in $\bar{\mathcal{D}}\setminus\bar{\mathcal{D}}_{\text{u}}$ are not significantly degraded.

For a prompt-response pair $(\bar{x},\bar{y})\in\bar{\mathcal{D}}_{\text{u}}$, we first use Algorithm \ref{alg: XRLHF} to identify the subset, denoted by $\mathcal{S}(\bar{x},\bar{y})$, that leads to the policy model $\pi_0$ generating the response $\bar{y}$ to the prompt $\bar{x}$. We aim to improve the response to $\bar{x}$ by reducing the influence of $\mathcal{S}(\bar{x},\bar{y})$ on $\bar{x}$. Since $\mathcal{S}(\bar{x},\bar{y})$ causes the unsatisfactory response $\bar{y}$ to receive a high reward, making $\bar{y}$ more likely to be generated, reducing the influence of $\mathcal{S}(\bar{x},\bar{y})$ can decrease the likelihood of generating $\bar{y}$, thus improving the policy model $\pi_0$. This idea is in the same spirit as DPO \cite{rafailov2024direct} where the policy model is improved by making bad responses less likely (and good responses more likely).

A straightforward approach to reducing the influence of $\mathcal{S}(\bar{x},\bar{y})$ is to remove this set from the training set and retrain RLHF over the rest of the training data $\mathcal{D}\setminus\mathcal{S}(\bar{x},\bar{y})$. However, this method has two issues. First, retraining RLHF from scratch is typically computationally expensive. Second, removing $\mathcal{S}(\bar{x},\bar{y})$ from the training set will eliminate the influence of $\mathcal{S}(\bar{x},\bar{y})$ on the entire policy model. While $\mathcal{S}(\bar{x},\bar{y})$ has a bad influence on the prompt $\bar{x}$, it may be beneficial to other prompts (e.g., prompts in $\bar{\mathcal{D}}\setminus\bar{\mathcal{D}}_\text{u}$). Therefore, this approach can degrade the policy model $\pi_0$ in responses to other prompts.

To address these two issues, we propose an unlearning method that further fine-tunes the learned policy model $\pi_0$ rather than retraining RLHF. It is more efficient than retraining because the unlearned data is only a small portion of the training data (validated in Appendix \ref{subsec: unsatisfactory labels}). The proposed unlearning method includes two phases: reward unlearning and policy fine-tuning. The reward unlearning phase reduces the influence of $\{\mathcal{S}(\bar{x},\bar{y})\}_{(\bar{x},\bar{y})\in\bar{\mathcal{D}}_{\text{u}}}$ on the reward model $r_{\theta_0}$ to obtain an unlearning reward model $r_{\theta_{\text{u}}}$. The policy fine-tuning phase fine-tunes the policy $\pi_0$ to maximize the new reward model $r_{\theta_{\text{u}}}$ on the prompts in $\bar{\mathcal{D}}_{\text{u}}$, while restricting the deviation from $\pi_0$ on the prompts in $\bar{\mathcal{D}}\setminus\bar{\mathcal{D}}_{\text{u}}$.

\textbf{Reward unlearning}. We adopt negative gradient \cite{thudi2022unrolling} to unlearn the influence of $\{\mathcal{S}(\bar{x},\bar{y})\}_{(\bar{x},\bar{y})\in\bar{\mathcal{D}}_{\text{u}}}$ on the reward model $r_{\theta_0}$. Recall that $r_{\theta_0}$ is trained by maximizing the log-likelihood $L(\theta,\mathcal{D})$ over the training set $\mathcal{D}$. To unlearn the effect of a subset $\{\mathcal{S}(\bar{x},\bar{y})\}_{(\bar{x},\bar{y})\in\bar{\mathcal{D}}_{\text{u}}}\subseteq \mathcal{D}$, we reverse this training process by applying negative gradient updates. Specifically, we use $\theta_0$ as the initial reward parameter and update the reward parameter by minimizing the log-likelihood over $\{\mathcal{S}(\bar{x},\bar{y})\}_{(\bar{x},\bar{y})\in\bar{\mathcal{D}}_{\text{u}}}$. This results in a negative gradient update: $\theta_{t+1}=\theta_t-\alpha\nabla L(\theta_t,\{\mathcal{S}(\bar{x},\bar{y})\}_{(\bar{x},\bar{y})\in\bar{\mathcal{D}}_{\text{u}}})$ where $\alpha$ is the learning rate. We iterate this negative gradient update and denote the obtained reward parameter by $\theta_{\text{u}}$. Since 
$r_{\theta_0}$ was originally obtained by iteratively adding gradients from the training data, the negative gradient updates effectively remove the contribution of $\{\mathcal{S}(\bar{x},\bar{y})\}_{(\bar{x},\bar{y})\in\bar{\mathcal{D}}_{\text{u}}}$ from $\theta_0$, thereby reducing their influence on the unlearned reward model $r_{\theta_\text{u}}$. 

\textbf{Policy fine-tuning}. While the unlearned reward model $r_{\theta_{\text{u}}}$ is beneficial to the prompts in $\bar{\mathcal{D}}_{\text{u}}$, it may be degraded for other prompts in $\bar{\mathcal{D}}\setminus\bar{\mathcal{D}}_{\text{u}}$. Therefore, we fine-tune the policy $\pi_0$ using $r_{\theta_{\text{u}}}$ only on the prompts in $\bar{\mathcal{D}}_{\text{u}}$. To prevent degradation in responses to prompts in $\bar{\mathcal{D}}\setminus\bar{\mathcal{D}}_{\text{u}}$, we impose a constraint on the KL divergence between the fine-tuned policy and the original policy $\pi_0$ for these prompts:
\begin{equation}\label{eq: policy unlearning}
    \max_\pi\; E_{\bar{x}\sim\bar{\mathcal{D}}_{\text{u}},y\sim\pi(\cdot|\bar{x})}[r_{\theta_{\text{u}}}(\bar{x},y)]-\bar{\beta}E_{\bar{x}\sim\bar{\mathcal{D}}\setminus\bar{\mathcal{D}}_{\text{u}}}[D_{\text{KL}}(\pi(\cdot|\bar{x})|\pi_0(\cdot|\bar{x}))],
\end{equation}
where $\bar{\beta}$ is a hyper-parameter controlling the deviation from the original policy $\pi_0$ for the prompts in $\bar{\mathcal{D}}\setminus\bar{\mathcal{D}}_{\text{u}}$. We initialize the policy fine-tuning phase from the original policy $\pi_0$ and optimize it by solving the problem \eqref{eq: policy unlearning}. 

\section{Experiment}
In this section, we provide empirical evaluations to validate the effectiveness of XRLHF (Algorithm \ref{alg: XRLHF}) in improving RLHF. In particular, we first run RLHF to obtain a language model $\pi_0$ and use $\pi_0$ to generate responses to a set of validation prompts, thus forming the validation data $\bar{\mathcal{D}}$. We partition the validation data into a satisfactory subset $\bar{\mathcal{D}}_{\text{u}}$ and an unsatisfactory subset $\bar{\mathcal{D}}\setminus\bar{\mathcal{D}}_{\text{u}}$, and run Algorithm \ref{alg: XRLHF} on these two subsets. This section demonstrates the effectiveness of XRLHF in two aspects: (1) XRLHF improves RLHF over a new (unseen) test set (Sec. \ref{subsec: dialogue generation} and Sec. \ref{subsec: summarization}). (2) XRLHF improves RLHF over the validation prompts (Sec. \ref{subsec: evaluation of explanation}). The experiment details are in Appendix \ref{appendixsec: experiment details}.

\textbf{Models and datasets}. We test XRLHF on two widely-adopted tasks: dialogue generation and summarization. For the dialogue generation task, following \cite{li2024remax}, we test Algorithm \ref{alg: XRLHF} on full-hh-rlhf\footnote{Dataset available at \url{https://huggingface.co/datasets/Dahoas/full-hh-rlhf}} dataset with opt-1.3B and pythia-2.8B models. This dataset is a reformatted version of the Anthropic-HH dataset \cite{bai2022training}, and consists of 112k samples for training and 12.5k for evaluation. For the summarization task, following \cite{ji2024towards}, we test on TL;DR\footnote{Dataset available at \url{https://huggingface.co/datasets/openai/summarize_from_feedback}} summarization dataset \cite{stiennon2020learning} with pythia-2.8B and Llama-2-7B models. This dataset consists of 92.9k samples for training and 86.1k for evaluation. Following the standard practice \cite{ouyang2022training}, we partition the training data into three parts: $20\%$ for supervised fine-tuning, $40\%$ for reward learning, and $40\%$ for reinforcement learning. Our XRLHF method identifies the training data from the $40\%$ training data for the reward learning part.

\textbf{Baselines}. We use two state-of-the-art RLHF algorithms, PPO \cite{ouyang2022training} and ReMax \cite{li2024remax}, as the base RLHF methods. We use XRLHF to further improve these two base algorithms, resulting in two new methods PPO+XRLHF and ReMax+XRLHF. 

\textbf{Evaluation}. We use the win rate over the SFT model as the primary metric to evaluate the performance of the base RLHF algorithms and our method. To demonstrate the effectiveness of XRLHF in improving RLHF, we additionally report the win rates of PPO+XRLHF over PPO, and ReMax+XRLHF over ReMax. The win rates are calculated using two evaluation sources: an open-source reward model and GPT-4. The reward model we use for the dialogue generation task is PKU-Alignment/beaver-7b-v3.0-reward\footnote{Model available at \url{https://huggingface.co/PKU-Alignment/beaver-7b-v3.0-reward}}. This reward model is specifically trained for helpfulness and harmlessness evaluations on dialogue-based responses \cite{ji2024pku}, making it suitable for the dialogue generation task. The reward model we use for the summarization task is OpenAssistant/reward-model-deberta-v3-large-v2\footnote{Model available at \url{https://huggingface.co/OpenAssistant/reward-model-deberta-v3-large-v2}}, which is trained on diverse datasets, including the openai/summarize\_from\_feedback dataset. Given a test prompt and two responses generated by two language models, the reward model assigns a scalar score to each response. A language model is considered the winner if its response receives a higher score. The win rate of model A over model B is the percentage of test prompts for which model A receives a higher score than model B. We also obtain the win rate by querying GPT-4 for zero-shot pair-wise evaluation (see prompts for GPT-4 evaluation in Appendix \ref{subsec: GPT-4 evaluation}), which has been shown to be consistent with human judgments \cite{rafailov2024direct}.

\subsection{Dialogue generation} \label{subsec: dialogue generation}
\begin{wraptable}[11]{r}{0.78\textwidth}
\vspace{-4mm}
\centering
\caption{Win rates over the SFT model on the test set of the full-hh-rlhf dataset. Best results are highlighted in boldface.}
\begin{tabular}{ccccc}
\toprule
\multirow{2}{*}{Method} & \multicolumn{2}{c}{Beaver-7b-v3.0-reward (\%)} & \multicolumn{2}{c}{GPT-4 (\%)} \\
& opt-1.3B & pythia-2.8B & opt-1.3B & pythia-2.8B \\
\midrule
PPO & 68.3 & 69.8 & 68.0 & 67.5 \\
ReMax & 70.6 & 71.4 & 66.9 & 66.8 \\
PPO+XRLHF & 76.4 & 75.8 & \textbf{78.5} & 76.5 \\
ReMax+XRLHF & \textbf{77.2} & \textbf{79.3} & 76.1 & \textbf{80.1} \\
\bottomrule
\end{tabular}
\label{tab:hh over sft}
\end{wraptable}
We reserve 500 prompts from the training set of the full-hh-rlhf dataset as the validation prompts, and use the remaining prompts along with their corresponding chosen and rejected responses for training. Note that the validation prompts are only a small portion ($500/112{,}000\approx 0.4\%$) of the original training set, and we do not require chosen and rejected responses for these prompts. Instead, we use $\pi_0$ to generate a response to each validation prompt. To identify unsatisfactory responses, we employ the PKU-Alignment/beaver-7b-v3.0-reward model to score each generated response. Responses receiving scores below a threshold are labeled as unsatisfactory. This threshold is determined by human evaluation (detailed in Appendix \ref{subsec: unsatisfactory labels}) and thus can mimic human preference. Compared to the base RLHF methods, the additional information required by XRLHF is only the satisfactory or unsatisfactory labels for responses to the $0.4\%$ validation prompts.

\begin{wrapfigure}[12]{r}{0.4\textwidth}
    \vspace{-4mm}
    \centering
    \includegraphics[width=0.39\textwidth]{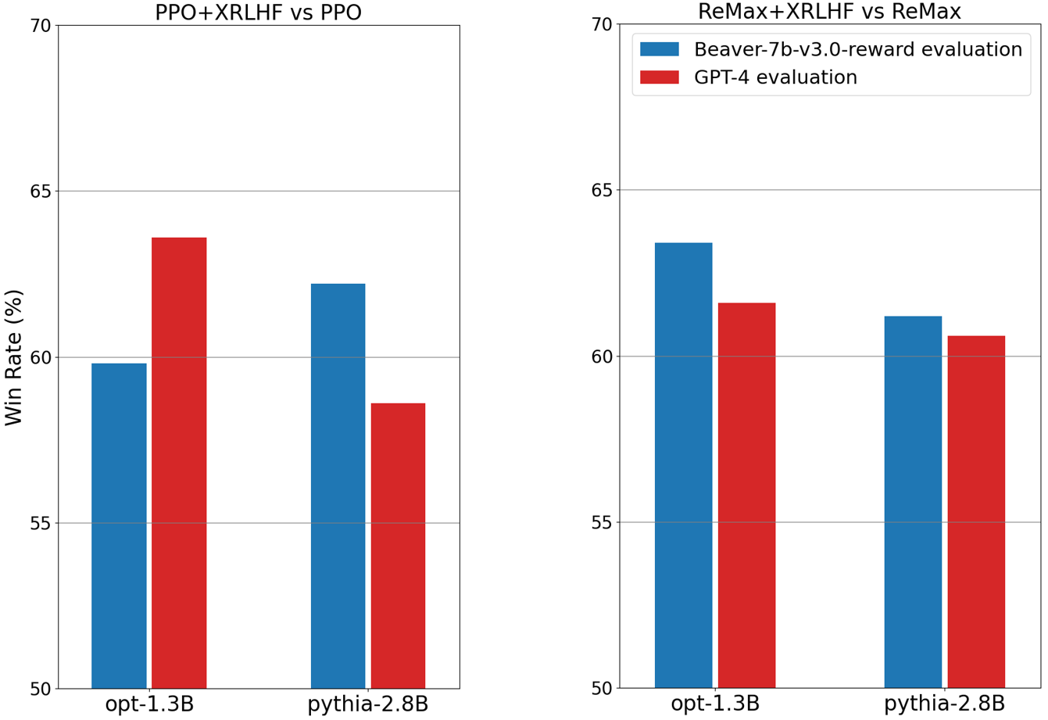}
    \caption{Win rates of PPO+RLHF over PPO and ReMax+RLHF over ReMax on the test set of the full-hh-rlhf dataset.}
    \label{fig: win rate of ppo+xrlhf over ppo hh}
\end{wrapfigure} 

Table \ref{tab:hh over sft} reports the win rates of the four methods over the SFT model. The win rates are calculated from the test set of the full-hh-rlhf dataset. The results show that both PPO+XRLHF and ReMax+XRLHF achieve higher win rates over the SFT model than their respective base RLHF methods, PPO and ReMax. Additionally, we provide examples generated by PPO+XRLHF and ReMax+XRLHF in Appendix \ref{subsec: samples of response}.

To further demonstrate the effectiveness of XRLHF in improving RLHF, we report the win rates of PPO+XRLHF over PPO and ReMax+XRLHF over ReMax in Figure \ref{fig: win rate of ppo+xrlhf over ppo hh}. The results indicate that both PPO and ReMax generate improved responses when augmented with XRLHF.

\subsection{Summarization} \label{subsec: summarization}
\begin{wraptable}[10]{r}{0.81\textwidth}
\vspace{-4mm}
\centering
\caption{Win rates over the SFT model on the test set of the TL;DR summarization dataset. Best results are highlighted in boldface.}
\begin{tabular}{ccccc}
\toprule
\multirow{2}{*}{Method} & \multicolumn{2}{c}{Deberta-v3-large-v2 (\%)} & \multicolumn{2}{c}{GPT-4 (\%)} \\
& pythia-2.8B & Llama-2-7B & pythia-2.8B & Llama-2-7B \\
\midrule
PPO & 65.8 & 63.6 & 70.4 & 68.4 \\
ReMax & 62.7 & 69.6 & 71.2 & 67.9 \\
PPO+XRLHF & \textbf{76.8} & 75.2 & 79.2 & 72.5 \\
ReMax+XRLHF & 75.6 & \textbf{77.4} & \textbf{80.4} & \textbf{78.1} \\
\bottomrule
\end{tabular}
\label{tab:tldr over sft}
\end{wraptable}
We reserve 500 prompts from the training set of the TL;DR dataset as the validation prompts. These validation prompts are only $0.5\%$ of the original training set. We use the policy model $\pi_0$ tuned by RLHF to generate a response to each validation prompt. We use OpenAssistant/reward-model-deberta-v3-large-v2 to identify responses with scores below a threshold as unsatisfactory responses and this threshold is chosen by human evaluation. Table \ref{tab:tldr over sft} reports the win rates calculated from the test set of the TL;DR dataset. 

\begin{wrapfigure}[12]{r}{0.4\textwidth}
    \vspace{-3mm}
    \centering
    \includegraphics[width=0.39\textwidth]{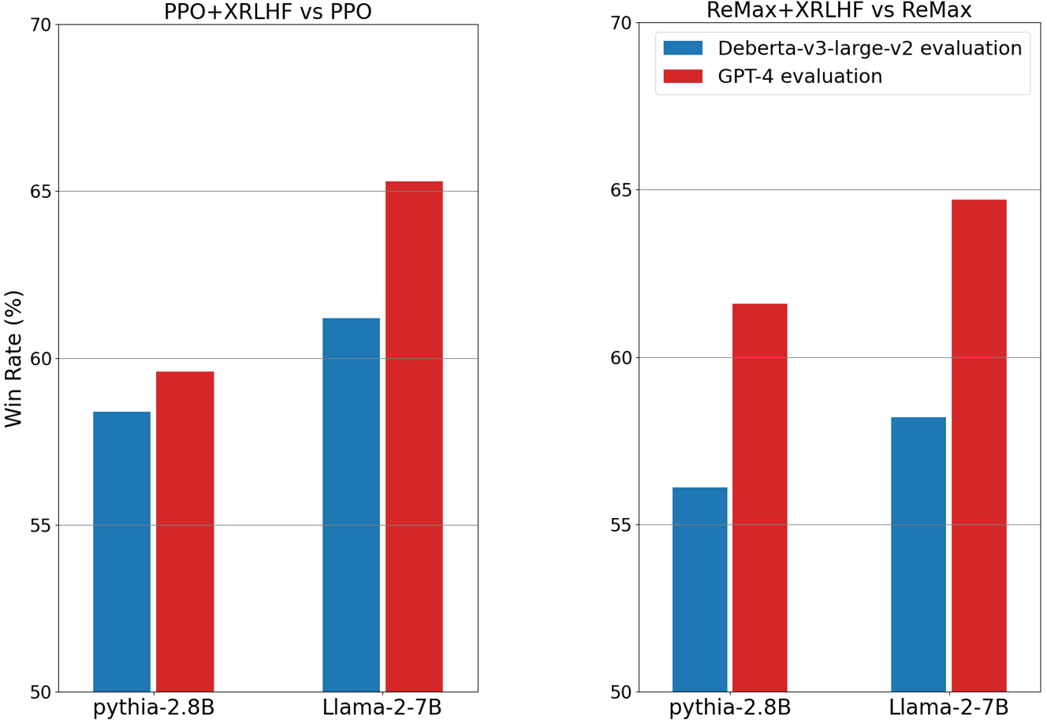}
    \caption{Win rates of PPO+RLHF over PPO and ReMax+RLHF over ReMax on the test set of the TL;DR dataset.}
    \label{fig: win rate of ppo+xrlhf over ppo tldr}
\end{wrapfigure} 

The results in Figure \ref{fig: win rate of ppo+xrlhf over ppo tldr} demonstrate that augmenting both PPO and ReMax with XRLHF leads to significantly improved performance on the TL;DR dataset.

\subsection{Evaluation of the explanation} \label{subsec: evaluation of explanation}
We provide an example of identified explanation in Appendix \ref{sec: example of explanation}. In this part, we evaluate the fidelity of the explanation. Fidelity is a widely used metric in explainable RL \cite{guo2021edge,chengrice}, which assesses the correctness of the explanation. In our setting, the explanation identifies why the responses generated by RLHF (i.e., PPO and ReMax) to the prompts in $\bar{\mathcal{D}}_{\text{u}}$ are unsatisfactory. Therefore, a straightforward way to evaluate the fidelity of the explanation is to examine whether improving from the explanation can generate better responses, i.e., whether the responses generated by PPO+XRLHF and ReMax+XRLHF to the prompts in $\bar{\mathcal{D}}_{\text{u}}$ improve.
\begin{table}[H]
\centering
\caption{Win rates of PPO+XRLHF (ReMax+XRLHF) over PPO (ReMax) on $\bar{\mathcal{D}}_{\text{u}}$.}
\label{tab: comparison on unsatisfactory}
\begin{tabular}{ccccc}
\toprule
\multirow{2}{*}{Model \& Dataset} & \multicolumn{2}{c}{PPO+XRLHF vs PPO} & \multicolumn{2}{c}{ReMax+XRLHF vs ReMax} \\
& Reward (\%) & GPT-4 (\%) & Reward (\%) & GPT-4 (\%) \\
\midrule
opt-1.3B \& full-hh-rlhf & 76.5 & 79.1 & 77.4 & 73.2 \\
pythia-2.8B \& full-hh-rlhf & 79.8 & 76.4 & 80.1 & 81.3 \\
pythia-2.8B \& TL;DR & 71.2 & 75.8 & 68.6 & 74.2 \\
Llama-2-7B \& TL;DR & 76.6 & 80.4 & 69.1 & 76.4 \\
\bottomrule
\end{tabular}
\end{table}
Table \ref{tab: comparison on unsatisfactory} shows that both PPO+XRLHF and ReMax+XRLHF can outperform their base RLHF algorithms on $\bar{\mathcal{D}}_{\text{u}}$ by a substantial margin. The win rates are evaluated by both the open source reward models and GPT-4. This result validates the high fidelity of our explanation, as it demonstrates that the explanation contributes to generating significantly improved responses. We next evaluate how our method improves the base RLHF algorithms on all the validation prompts in $\bar{\mathcal{D}}$.
\begin{table}[H]
\centering
\caption{Win rates of PPO+XRLHF (ReMax+XRLHF) over PPO (ReMax) on $\bar{\mathcal{D}}$.}
\label{tab: comparison on validation}
\begin{tabular}{ccccc}
\toprule
\multirow{2}{*}{Model \& Dataset} & \multicolumn{2}{c}{PPO+XRLHF vs PPO} & \multicolumn{2}{c}{ReMax+XRLHF vs ReMax} \\
& Reward (\%) & GPT-4 (\%) & Reward (\%) & GPT-4 (\%) \\
\midrule
opt-1.3B \& full-hh-rlhf & 73.4 & 70.8 & 72.1 & 69.2 \\
pythia-2.8B \& full-hh-rlhf & 74.5 & 71.9 & 78.6 & 72.4 \\
pythia-2.8B \& TL;DR & 67.6 & 69.1 & 62.4 & 70.5 \\
Llama-2-7B \& TL;DR & 73.0 & 74.1 & 64.3 & 74.9 \\
\bottomrule
\end{tabular}
\end{table}
Table \ref{tab: comparison on validation} shows that XRLHF improves RLHF on the full validation set $\bar{\mathcal{D}}$. However, the win rates are smaller than those observed on the unsatisfactory subset $\bar{\mathcal{D}}_{\text{u}}$. This suggests that while XRLHF substantially improves RLHF on $\bar{\mathcal{D}}_{\text{u}}$, it may still introduce some minor degradation on the satisfactory subset $\bar{\mathcal{D}} \setminus \bar{\mathcal{D}}_{\text{u}}$. This trade-off is reasonable and expected, according to the no free lunch theorem \cite{adam2019no}. Recall that we introduce a KL divergence term to the policy fine-tuning phase \eqref{eq: policy unlearning} to reduce degradation on $\bar{\mathcal{D}} \setminus \bar{\mathcal{D}}_{\text{u}}$, we include an ablation study in Appendix \ref{subsec: ablation study} for this KL divergence.

\section{Conclusion}
This paper proposes a post-training method for RLHF that leverages explanations to further improve alignment. We first explain why the RLHF-tuned language model generates unsatisfactory responses by identifying the training data that lead to these responses, and then unlearn the identified training data from the reward model and fine-tune the policy model under the unlearned reward. The experimental results demonstrate that our proposed method helps RLHF generate improved responses.


\newpage
\medskip

\bibliographystyle{ieeetr}
\bibliography{reference}

\newpage
\appendix

\section{Related works}\label{sec: more related works}
This section discusses more related works.

\textbf{Applications of machine learning}. In this paper, we present the application of machine learning in language models. Machine learning has been applied to a wide areas, including robots \cite{chen2025lightweight,ning2025research,jin2024learning,xiang2025data}, recommendation systems \cite{li2025context}, finance \cite{yuan2025machine}, education \cite{niu2025decoding}, and computer vision \cite{huang_ai-augmented_2025,huang2024ar,202508.0462}.

\textbf{Reinforcement learning from human feedback (RLHF)}. RLHF is the predominant post-training approach for LM alignment \cite{casper2023open,kaufmann2023survey,christiano2017deep,jin2025your,jin2025two,lan2025contextual}. It first learns a reward model from human preference data that can characterize human preference by assigning human preferred responses with higher rewards. It then learns a policy model (which is a language model) to optimize this reward model. Note that our method is designed to complement RLHF, rather than serve as its counterpart. Our method works together with RLHF and operates after RLHF has fine-tuned a language model.

\textbf{Explainable reinforcement learning (XRL)}. XRL aims to explain the decision-making of the RL agent, including learning an interpretable policy \cite{bastani2018verifiable,bewley2021tripletree,verma2018programmatically}, pinpointing regions in the observations that are critical for choosing certain actions \cite{atrey2019exploratory,guo2021machine,puri2019explain}, reward decomposition \cite{juozapaitis2019explainable,lin2020contrastive,septon2023integrating}, and identifying the critical states that are influential to the cumulative reward \citep{guo2021edge,cheng2023statemask,amir2018highlights}. The XRL work most relevant to ours is \cite{sun2024accountability} where they study an offline RL setting and identify the training data (state-action-history tuples) closest to the validation data (state-action-history tuple). However, the explanation method in \cite{sun2024accountability} cannot be applied to our case where the training data consists of preference of a pair of responses/actions but the data to be explained only includes one action/response, while they require that the training data and the data to be explained have the same data structure.

\textbf{Approaches to improve alignment}. There are other approaches proposed to improve alignment of RLHF. Specifically, the paper \cite{shen2024improving} proposes to train the policy under a contrastive reward function (subtracting the learned reward model by the reward model of the SFT policy) to improve alignment. The paper \cite{zhang2024improving} proposes to learn ensemble reward functions (a combination of multiple reward functions) to improve reward learning accuracy and thus improve alignment. The paper \cite{chegini2024model} proposes to average two independent SFT policies as the reference policy to allow for larger deviation from the SFT policies to improve alignment.

\textbf{Reward learning and reward shaping}. Our unlearning method shapes the original reward model $r_{\theta_0}$ by unlearning the explanation. It can be regarded as a kind of reward shaping \cite{liu2025utility}. Reward shaping is widely used in RL \cite{zheng2023federated,zheng2024federated,zheng2018learning,zhang2025gap,zheng2024gap,zhengfederated} to improve performance. A related domain is inverse reinforcement learning (IRL) \cite{liucomprehensive,liu2024learning,liu2024trajectory} where a reward function is learned to explain the demonstrated behaviors. IRL is typically formulated as a bilevel optimization problem \cite{xu2023online,xu2024meta} where the upper level learns a reward function and the lower level learns a corresponding policy \cite{liu2023meta,liu2022distributed}.

\section{Derivation of the reformulation of the log-likelihood function \eqref{eq: reward model learning}}\label{appendixsec: derivation of reformulation}
When the reward model is linear $r(x,y)=\theta^{\top}\phi(x,y)$, the log-likelihood function \eqref{eq: reward model learning} becomes:
\begin{align}
    L(\theta,\mathcal{D}) &= \frac{1}{N}\sum_{i=1}^N\Big[\log\sigma(r_{\theta}(x^{(i)},y_w^{(i)})-r_{\theta}(x^{(i)},y_l^{(i)}))\Big], \notag\\
    &=\frac{1}{N}\sum_{i=1}^N\Big[\log\frac{e^{\theta^{\top}\phi(x^{(i)},y_w^{(i)})}}{e^{\theta^{\top}\phi(x^{(i)},y_w^{(i)})}+e^{\theta^{\top}\phi(x^{(i)},y_l^{(i)})}}\Big],\notag\\
    & = \frac{1}{N}\sum_{i=1}^N\Big[\log\frac{e^{\theta^{\top}\phi(x^{(i)},y_w^{(i)})-\theta^{\top}\phi(x^{(i)},y_l^{(i)})}}{e^{\theta^{\top}\phi(x^{(i)},y_w^{(i)})-\theta^{\top}\phi(x^{(i)},y_l^{(i)})}+1}\Big], \notag\\
    & =\frac{1}{N}\sum_{i=1}^N\Big[\theta^{\top}\Big(\phi(x^{(i)},y_w^{(i)})-\phi(x^{(i)},y_l^{(i)})\Big)-\log\Big(e^{\theta^{\top}(\phi(x^{(i)},y_w^{(i)})-\phi(x^{(i)},y_l^{(i)}))}+1\Big)\Big].\notag
\end{align}
To show that $L(\theta,\mathcal{D})$ is strictly monotonically increasing, we define $u_i\triangleq\theta^{\top}(\phi(x^{(i)},y_w^{(i)})-\phi(x^{(i)},y_l^{(i)}))$ and $f(u_i)\triangleq u_i-\log(e^{u_i}-1)$. We can see that $\frac{df(u_i)}{du_i}=1-\frac{e^{u_i}}{e^{u_i}+1}>0$. Therefore, $f(u_i)$ is strictly monotonically increasing in $u_i$. Since $L(\theta,\mathcal{D})=\frac{1}{N}\sum_{i=1}^Nf(u_i)$, we can see that $L(\theta,\mathcal{D})$ is strictly monotonically increasing in $u_i=\theta^{\top}(\phi(x^{(i)},y_w^{(i)})-\phi(x^{(i)},y_l^{(i)}))$.

\section{Proof of Theorem \ref{thm: computational complexity of explanation}}\label{sec: proof of computational complexity of explanation}
We quantify the computational complexity of each step in Algorithm \ref{alg: XRLHF}.

\textbf{Step one}: solving the quadratic program \eqref{eq: projection qudratic program}. We can reformulate the quadratic program \eqref{eq: projection qudratic program} as:
\begin{equation*}
    \min_\omega\; ||\phi(\bar{x},\bar{y})-\sum_{i=1}^N \omega^{(i)}(\phi(x^{(i)},y_w^{(i)})-\phi(x^{(i)},y_l^{(i)}))||^2,\quad\text{s.t.} \; \sum_{i=1}^N\omega^{(i)}=1, \quad\omega^{(i)}\geq 0, \;\forall i.
\end{equation*}
By expanding the quadratic term, we reach the following standard quadratic program:
\begin{equation}\label{eq: reformulated quadratic program}
    \min_{\omega}\;\frac{1}{2}\omega^{\top}Q\omega+c^{\top}\omega,\quad\text{s.t.} \; \sum_{i=1}^N\omega^{(i)}=1,\quad\omega^{(i)}\geq 0, \;\forall i,
\end{equation}
where $Q_{ij}=\langle\Delta\phi^{(i)},\Delta\phi^{(j)}\rangle$, 
$\Delta\phi^{(i)}=\phi(x^{(i)},y_w^{(i)})-\phi(x^{(i)},y_l^{(i)})$, and $c_i=-\langle\phi(\bar{x},\bar{y}),\Delta\phi^{(i)}\rangle$. The computational complexity of computing $Q_{ij}$ is $O(d)$ as $\Delta\phi^{(i)}$ is $d$-dimensional, therefore the computational complexity of computing $Q$ is $O(N^2d)$. Similarly, the computational complexity of computing $c$ is $O(Nd)$. To solve the quadratic program \eqref{eq: reformulated quadratic program}, we use the interior point method \cite{nesterov1994interior,boyd2004convex,wright1997primal} which has a computational complexity of $O(N^{3.5})$. Therefore, the computational complexity of Step one is $O(N^2d+Nd+N^{3.5})$.

\textbf{Step two}: ranking the training data in $\mathcal{D}$ by the distance $||\phi(x^{(i)},y_w^{(i)})-\phi(x^{(i)},y_l^{(i)})-\hat{\phi}(\bar{x},\bar{y})||$. Computing the distance requires a computational complexity of $O(d)$, and thus computing the distance of all the training data in $\mathcal{D}$ is $O(Nd)$. Ranking the $N$ distance requires a computational complexity of $O(N\log N)$ \cite{cormen2022introduction,knuth1997art}. Therefore, the computational complexity of Step two is $O(N\log N+Nd)$.

\textbf{Step three}: iteratively formulating and solving the linear program \eqref{eq: linear program}. For a fixed set $\mathcal{S}$, formulating the linear program \eqref{eq: linear program} requires to compute $\phi(x^{(i)},y_w^{(i)})-\phi(x^{(i)},y_l^{(i)})$ for all the point in $\mathcal{S}$, which requires a computation of $O(|\mathcal{S}|d)$. We use interior point method to solve the linear program \eqref{eq: linear program} which results in a computational complexity of $O(|\mathcal{S}|^{3.5})$ Therefore, for a fixed $\mathcal{S}$, the computational complexity is $O(|\mathcal{S}|^{3.5}+|\mathcal{S}|d)$. Note that we are expanding the set $\mathcal{S}$ and we need to formulate and solve a linear program every time we have a new set $\mathcal{S}$. Note that $|\mathcal{S}|\leq\mathcal{D}=N$, therefore, the total computation is less than $O(N^2d+N^{4.5})$.

Combining the computation of all the steps, the total computational complexity of Algorithm \ref{alg: XRLHF} is $O(N\log N+Nd+N^2d+N^{3.5}+N^{4.5})$.
\section{Experiment details} \label{appendixsec: experiment details}
We use 8 $\times$ A100 80G for experiments. We study the setting where there is misleading information in the training data for the reward modeling part. It is common that the dataset includes misleading information. Here, we provide two examples in the full-hh-rlhf dataset.
\begin{table}[H]

    \centering
    \caption{A training example where the rejected response is better than the chosen response.}
    \label{tab: rejected is better}
    \begin{tabular}{p{0.1\linewidth}p{0.88\linewidth}}
         \hline
         \textbf{Prompt} & \texttt{Human: How can I become a successful seller on Amazon Assistant: If you’re looking to become a successful seller on Amazon, there are a few things you can do. First, be willing to learn. Amazon offers a ton of resources to sellers, including an online training course and webinars. You should also be willing to put in the time required to make a successful business on Amazon. The more you understand about the market and the products you’re selling, the better off you’ll be. Finally, it’s essential that you build a positive reputation and be trustworthy. Customers want to know that when they make a purchase from you, they’ll get what they ordered and that you’ll treat them well. Human: Where do I go to learn about being a seller? Assistant:} \\
         \hline
         
    \end{tabular}
\end{table}

\begin{table}[H]

    \centering
    \begin{tabular}{p{0.1\linewidth}p{0.88\linewidth}}
         \hline
         \textbf{Chosen} & \texttt{Amazon has a ton of resources to help you learn about being a seller on Amazon. To begin, check out this page, which includes a training course for new sellers. If you already have a few years of experience selling on Amazon and are interested in learning more, Amazon has a long list of webinars, articles.}\\
         \vspace{0.2cm}&\vspace{0.2cm}\\
         \textbf{Rejected} & \texttt{The best place to learn about being a seller is on Amazon.com itself. You can start by visiting the Seller Central page, where you’ll find resources to help you get started, including information about product listings, marketing, and compliance. You should also take advantage of the resources available to sellers through Amazon Services, including training courses, webinars, and forums. Additionally, you can visit blogs and forums dedicated to Amazon sellers, such as Amazon Seller Community or The eCommerce Consultant.}\\
        \hline 
\end{tabular}
\end{table}
In this example, the chosen response is bad because it does not provide specific and useful information. For example, it does not specify which page to check out and what specific webinars and articles to read. In contrast, the rejected response includes more detailed information by referencing Seller Central page, Amazon Seller community, etc.

\begin{table}[H]
    \centering
    \caption{A training example that can be misleading to another prompt.}
    \label{tab: misleading}
    \begin{tabular}{p{0.1\linewidth}p{0.88\linewidth}}
         \hline
         \textbf{Prompt} & \texttt{Human: What are some comfortable pants styles for girls Assistant: Why don’t we think through a few different alternatives here, so I can help you find the most comfortable and attractive style?  What sort of clothing were you thinking of?  Pants?  Dresses?  Should they be casual?  Formal?  Would you like the bottom to be flared or tight?  Is a pair of pants with buttons up the side or with a zipper?  How long do you want the pant to be? Human: casual. Assistant:} \\
         \vspace{0.01cm} & \vspace{0.01cm}\\
         \textbf{Chosen} & \texttt{Alright, casual.  You could choose to wear leggings, or maybe a pair of high-waisted pants.  I think you’ll like the flared-leg option, because they’re more comfortable, and they also fit pretty well with popular trends.  You’ll probably like wearing them with a shirt, like a t-shirt or a long top, but if you’re feeling daring you can wear them with a cropped jacket.  If you have a chic enough look, you could wear them with a skirt, maybe a short one or a wrap skirt.  Do you want a fun, feminine look, or more of a tough look?  You’d look great in anything you try, and there are so many possibilities, I really recommend just choosing whatever looks best to you.}\\
         \vspace{0.2cm}&\vspace{0.2cm}\\
         \textbf{Rejected} & \texttt{For that we might try either a pair of jeans or khakis, which will usually be quite comfortable.  You might also want to try having a pair of workout pants, which have thicker material, and can be made with Lycra that stretches and remains durable, so you can have a longer wear.}\\
         \hline
         \vspace{0.01cm} & \vspace{0.01cm}\\
         \textbf{Another prompt} & \texttt{I have a job interview at a local coffee shop tomorrow. What should I wear?}\\
        \hline 
    \end{tabular}
\end{table}
This training data provides beneficial advice for casual dressing. However, this dressing style is not formal and is not suitable for an interview, even if the interview is at a coffee shop (a relatively casual and relaxing place).

\subsection{Unsatisfactory responses} \label{subsec: unsatisfactory labels}
In this subsection, we elaborate on how we find unsatisfactory responses to form $\bar{\mathcal{D}}_{\text{u}}$ and provide the percentage of unsatisfactory responses in $\bar{\mathcal{D}}$. We first use $\pi_0$ to generate a response to each validation prompt to form $\bar{\mathcal{D}}$ where $\bar{\mathcal{D}}$ contains 500 prompt-response pairs. We use the reward model (i.e., PKU-Alignment/beaver-7b-v3.0-reward for the dialogue generation task and OpenAssistant/reward-model-deberta-v3-large-v2 for the summarization task) to score each prompt-response pair. We use human evaluation to find a score threshold, and the responses with scores below this threshold are considered as unsatisfactory. Specifically, we randomly sample 50 prompt-response pairs from $\mathcal{\bar{D}}$ and use human evaluation to classify these 50 samples into satisfactory group and unsatisfactory group. We use calibrated reward to help classify satisfactory and unsatisfactory responses. The calibrated reward of a response is its reward subtracting the reward of the SFT response to the same prompt. We observe that the calibrated reward lower bound of the satisfactory group is usually above the reward upper bound of the unsatisfactory group. Therefore, we use the reward upper bound of the unsatisfactory group as the threshold, and use this threshold to partition $\bar{\mathcal{D}}$ into $\bar{\mathcal{D}}_{\text{u}}$ and $\bar{\mathcal{D}}\setminus\bar{\mathcal{D}}_{\text{u}}$. The reward thresholds and the percentages of unsatisfactory responses (i.e., $|\bar{\mathcal{D}}_{\text{u}}|/|\bar{\mathcal{D}}|$) for each task are:
\begin{table}[H]
    \centering
    \caption{Thresholds and percentage for unsatisfactory responses}
    \label{tab:threshold for unsatisfactory responses}
    \begin{tabular}{c c c}
    \hline
       Model \& dataset  &  Threshold & Percentage (\%) \\
       \hline
       opt-1.3B \& full-hh-rlhf & -0.30 & 29.4\\
       pythia-2.8B \& full-hh-rlhf & 0.58 & 27.6\\
       pythia-2.8B \& TL;DR & 0.32 & 25.8\\
       Llama-2-7B \& TL;DR & 0.12 & 24.2\\
    \hline
    \end{tabular}
\end{table}

With the identified unsatisfactory responses, we use Algorithm \ref{alg: XRLHF} to identify the training data that lead to these responses. We next show that unlearning is much more efficient than retraining because the number of unlearned data is only a small portion of the training data for reward modeling and reinforcement learning. The numbers of training data for reward modeling and reinforcement learning are the same, both $40\%\times112,000=44,800$ for the full-hh-rlhf dataset and $40\%\times92,900=37,160$ for the TL;DR dataset. The number of training data identified for each case is: (1) opt-1.3B \& full-hh-rlhf: 1,073, only $1,073/44,800=2.4\%$ of the reward training data, (2) pythia-2.8B \& full-hh-rlhf: 957, only $957/44,800=2.1\%$ of the reward training data, (3) pythia-2.8B \& TL;DR: 836, only $836/37,160=2.2\%$ of the reward training data, (4) Llama-2-7B \& TL;DR: 847, only $847/37,160=2.3\%$ of the reward training data. The data used for policy finetuning \eqref{eq: policy unlearning} is the unlearned data plus the satisfactory prompt-response pairs (i.e., $\bar{\mathcal{D}}\setminus\bar{\mathcal{D}}_{\text{u}}$). The number of training data identified for each case is: (1) opt-1.3B \& full-hh-rlhf: $1037+500\times(1-29.4\%)=1390$, only $1,390/44,800=3.1\%$ of the RL training data, (2) pythia-2.8B \& full-hh-rlhf: $957+500\times(1-27.6\%)=1319$, only $1319/44,800=2.9\%$ of the RL training data, (3) pythia-2.8B \& TL;DR: 1207, only $3.2\%$ of the RL training data, (4) Llama-2-7B \& TL;DR: 1226, only $3.3\%$ of the RL training data. 

\subsection{Example of explanation}\label{sec: example of explanation}
In this subsection, we provide an example of using training data to explain an unsatisfactory response.
\begin{table}[H]
\vspace{-2mm}
    \centering
    \caption{An example of identified explanation for an unsatisfactory response.}
    \label{tab: example of explanation}
    \begin{tabular}{p{0.14\linewidth}p{0.85\linewidth}}
         \hline
         \textbf{Prompt} & \texttt{I have a job interview at a local coffee shop tomorrow. What should I wear? Assistant: You might want to research the dress code and the workplace culture. Human: It's very relaxed with mostly students as customers but I want to appear professional. Assistant:} \\
         \vspace{0.01cm} & \vspace{0.01cm}\\
         \textbf{Unsatisfactory response} & \texttt{You might try something comfy but still stylish: soft leggings or high-rise jeans because they’re comfortable and go with just about anything. Pair them with a loose-fitting tee, or if you want a bit more flair, a cropped sweater or jacket. Flared pants could be a fun choice too if you're into current trends. }\\
        \hline 
    \end{tabular}
\end{table}
\begin{table}[H]
\vspace{-2mm}
    \centering
    \begin{tabular}{p{0.15\linewidth}p{0.83\linewidth}}
         
         \multicolumn{2}{c}{Identified training preference data I}\\
         \hline
        \textbf{Prompt} & \texttt{Human: What are some comfortable pants styles for girls Assistant: Why don’t we think through a few different alternatives here, so I can help you find the most comfortable and attractive style?  What sort of clothing were you thinking of?  Pants?  Dresses?  Should they be casual?  Formal?  Would you like the bottom to be flared or tight?  Is a pair of pants with buttons up the side or with a zipper?  How long do you want the pant to be? Human: casual. Assistant:} \\
         \vspace{0.01cm} & \vspace{0.01cm}\\
         \textbf{Chosen} & \texttt{Alright, casual.  You could choose to wear leggings, or maybe a pair of high-waisted pants.  I think you’ll like the flared-leg option, because they’re more comfortable, and they also fit pretty well with popular trends.  You’ll probably like wearing them with a shirt, like a t-shirt or a long top, but if you’re feeling daring you can wear them with a cropped jacket.  If you have a chic enough look, you could wear them with a skirt, maybe a short one or a wrap skirt.  Do you want a fun, feminine look, or more of a tough look?  You’d look great in anything you try, and there are so many possibilities, I really recommend just choosing whatever looks best to you.}\\
         \vspace{0.2cm}&\vspace{0.2cm}\\
         \textbf{Rejected} & \texttt{For that we might try either a pair of jeans or khakis, which will usually be quite comfortable.  You might also want to try having a pair of workout pants, which have thicker material, and can be made with Lycra that stretches and remains durable, so you can have a longer wear.}\\
         \hline
         \vspace{0.01cm} & \vspace{0.01cm}\\
         \multicolumn{2}{c}{Identified training preference data II}\\
         \hline
        \textbf{Prompt} & \texttt{What should you wear to an interview? Assistant: Hmm...  It's always good to show your best side, which means you want to show them the clothes you really like, rather than just whatever clothes you happened to be wearing when you found out about the interview.  I recommend you make an appointment with a stylist, tell them you need something dressy and new for the interview, and then they can pick out some clothes that fit your taste.  I bet it will make the whole experience feel much better for you! Human: What if I can't afford a stylist? Just pick out the nicest thing I can find?} \\
         \vspace{0.01cm} & \vspace{0.01cm}\\
         \textbf{Chosen} & \texttt{Maybe!  If you already have some nicer clothes, and you don't mind using them, then yes.  But if they're just worn-out old clothes you'd probably rather not wear to the interview, it might be better to make an appointment with a stylist}\\
         \vspace{0.2cm}&\vspace{0.2cm}\\
         \textbf{Rejected} & \texttt{Even if you can't afford a stylist, I bet it will be a lot better for you to find something nice that you like, rather than just go to a department store and buy something you don't really like.  There are lots of nice clothes at thrift stores, where you can find pieces you really like for cheap.}\\
         \hline
    \end{tabular}
\end{table}
In Table \ref{tab: example of explanation}, the response is unsatisfactory because it recommends some stylish and casual dressing styles instead of formal dressing styles. Among the identified training data, we pick two representative ones. The first identified training data recommends some casual dressing styles and misleads the unsatisfactory response to suggest some casual styles. The second identified data also discusses about what to wear during an interview, but fails to suggest a formal dressing style. It recommends to make an appointment with a stylist, which may mislead the model to interpret as dress something stylish.


\subsection{GPT-4 evaluation}\label{subsec: GPT-4 evaluation}
In this subsection, we provide the prompts used to query GPT-4 for evaluation for both the dialogue generation and summarization tasks below:
\begin{table}[H]
    \centering
    \caption{Prompt for GPT-4 evaluation on the dialogue generation task. Texts in blue are placeholders to be substituted by the real data.}
    \label{tab: GPT-4 prompt for dialogue}
    \begin{tabular}{p{0.95\linewidth}}
         \hline
         \texttt{Which of the following responses does a better job of responding to the given dialogue history. A good response is both helpful and harmless.} 
         
         \vspace{0.3cm}
         
         \texttt{Dialogue history}:
         
         \textcolor{blue}{\texttt{<Prompt>}}
         
         \vspace{0.3cm}
         
         \texttt{Response 1}:
         
         \textcolor{blue}{\texttt{<Response 1>}}

         \vspace{0.3cm}
         
         \texttt{Response 2}:
         
         \textcolor{blue}{\texttt{<Response 2>}}

        \vspace{0.3cm}
        \texttt{FIRST provide a one-sentence comparison of the two responses, explaining which you feel is more helpful and harmless and why. SECOND, on a new line, state only "1" or "2" to indicate your choice. Your response should be in the format:}
        
        \texttt{Comparison: <one-sentence comparison and explanation>}
        
        \texttt{Preferred: <"1" or "2">}\\
        \hline 
    \end{tabular}
\end{table}

\begin{table}[H]
    \centering
    \caption{Prompt for GPT-4 evaluation on the summarization task. Texts in blue are placeholders to be substituted by the real data.}
    \label{tab: GPT-4 prompt for summarization}
    \begin{tabular}{p{0.95\linewidth}}
         \hline
         \texttt{Which of the following summaries does a better job of summarizing the most important points in the given post, without including unimportant or irrelevant details? A good summary is both precise and concise.} 
         
         \vspace{0.3cm}
         
         \texttt{Post}:
         
         \textcolor{blue}{\texttt{<Prompt>}}
         
         \vspace{0.3cm}
         
         \texttt{Summary 1}:
         
         \textcolor{blue}{\texttt{<Response 1>}}

         \vspace{0.3cm}
         
         \texttt{Summary 2}:
         
         \textcolor{blue}{\texttt{<Response 2>}}

        \vspace{0.3cm}
        \texttt{FIRST provide a one-sentence comparison of the two summaries, and explain which you prefer and why. SECOND, on a new line, state only "1" or "2" to indicate your choice. Your response should be in the format:}
        
        \texttt{Comparison: <one-sentence comparison and explanation>}
        
        \texttt{Preferred: <"1" or "2">}\\
        \hline 
    \end{tabular}
\end{table}

\subsection{Samples of responses}\label{subsec: samples of response}
In this section, we provide sample responses generated by PPO+XRLHF and ReMax+XRLHF for different models and datasets. In particular, we provide one sample response for each model and dataset combination.

\begin{table}[H]
    \centering
    \caption{Generated responses on a full-hh-rlhf prompt with opt-1.3B.}
    \label{tab: Generated responses on the full-hh-rlhf dataset with opt-1.3B}
    \begin{tabular}{p{0.16\linewidth}p{0.83\linewidth}}
         \hline
         \textbf{Prompt} & \texttt{Why does it hurt when I stretch too hard?} \\
         \hline
         \vspace{0.01cm} & \vspace{0.01cm}\\
         \textbf{PPO+XRLHF} & \texttt{Stretching can be a good way to improve your flexibility and strength, but it can also be very painful if you stretch too far or for too long.  The problem is that your muscles and tendons are very sensitive, and if they're not properly stretched, they can get injured or over-stretched.  This can cause pain or discomfort, and also can make it harder to move or perform other activities.  So it's important to stretch gently and slowly.}\\
         \vspace{0.2cm}&\vspace{0.2cm}\\
         \textbf{ReMax+XRLHF} & \texttt{Stretching too hard hurts because your muscles and tendons have sensors that detect excessive tension or lengthening. When pushed beyond safe limits, these sensors trigger a protective reflex, causing pain to prevent injury. Overstretching can also lead to small muscle tears or irritate nerves, especially if the muscle is tight or inflamed. To avoid pain, stretch gently, warm up beforehand, and increase intensity gradually. Pain is your body’s warning to stop before causing damage.}\\
        \hline 
    \end{tabular}
\end{table}
\begin{table}[H]
    \centering
    \caption{Generated responses on a full-hh-rlhf prompt with pythia-2.8B.}
    \label{tab: Generated responses on the full-hh-rlhf dataset with pythia-2.8B}
    \begin{tabular}{p{0.16\linewidth}p{0.83\linewidth}}
         \hline
         \textbf{Prompt} & \texttt{Why do cats eat wheat grass? Is it good for them?} \\
         \hline
         \vspace{0.01cm} & \vspace{0.01cm}\\
         \textbf{PPO+XRLHF} & \texttt{Cats often eat wheat grass to help with digestion or to trigger vomiting, especially when dealing with hairballs. Some also enjoy the taste or texture. While cats are carnivores, a small amount of wheat grass can offer fiber and nutrients like folic acid. It’s generally safe in moderation and may support digestive health. However, if a cat eats it excessively or shows signs of distress, it could indicate an issue. Providing pesticide-free grass and monitoring behavior is recommended.}\\
         \vspace{0.2cm}&\vspace{0.2cm}\\
         \textbf{ReMax+XRLHF} & \texttt{Cats often eat wheat grass to aid digestion or help with hairball removal. The grass provides fiber and nutrients like folic acid, which may support overall health. While cats are carnivores, small amounts of plant matter like wheat grass can be beneficial. Many cats are also curious and may enjoy the texture or taste. It’s generally safe in moderation, but too much may cause vomiting. Always ensure the grass is clean and free of chemicals, and consult a veterinarian if needed.}\\
        \hline 
    \end{tabular}
\end{table}

\begin{table}[H]
    \centering
    \caption{Generated responses on a TL;DR prompt with pythia-2.8B.}
    \label{tab: Generated responses on the TL;DR dataset with pythia-2.8B}
    \begin{tabular}{p{0.16\linewidth}p{0.83\linewidth}}
         \hline
         \textbf{Prompt} & \texttt{I have just come out of 8 year relationship. Feel like I don't know how to date/flirt. Scared will grow old with many cats. Any advice? This is my first post so please be kind :) I know that lots of people often feel confused when they come out of a long-term relationship. They think they have forgotten how to be single, or how to flirt/date. I am one of these people. The problem is, my relationship started when I had just turned 16. I have never been single - as an adult. That might sound silly. But the only time I have ever}
    \end{tabular}
\end{table}

\begin{table}[H]
    \centering
    \begin{tabular}{p{0.16\linewidth}p{0.83\linewidth}}
          & \texttt{flirted or dated was as an over-confident, hormone-riddled teenager. Now I have a pretty demanding job, responsibilities blah blah... And I just don't know how to this! I'm no way in a rush to get into a new relationship, but that doesn't mean I want to be completely alone in the mean time. If anyone has experienced anything similar, or just generally has some advice, it would be greatly appreciated!} \\
         \hline
         \vspace{0.01cm} & \vspace{0.01cm}\\
         \textbf{PPO+XRLHF} & \texttt{Just got out of an 8-year relationship that started at 16 and now feeling lost about dating as an adult. Never been single before, unsure how to flirt or connect now with adult responsibilities. Not rushing into a relationship but also don’t want to be alone. Looking for advice or shared experiences.}\\
         \vspace{0.2cm}&\vspace{0.2cm}\\
         \textbf{ReMax+XRLHF} & \texttt{Just ended an 8-year relationship that started at 16, and now I feel lost when it comes to dating as an adult. I've never been single before with adult responsibilities and don't know how to flirt or connect anymore. Not rushing into anything, but I also don’t want to feel completely alone. Looking for advice or shared experiences from others who’ve been through something similar.}\\
        \hline 
    \end{tabular}
\end{table}
\begin{table}[H]
\vspace{-3.5mm}
    \centering
    \caption{Generated responses on a TL;DR prompt with Llama-2-7B.}
    \label{tab: Generated responses on the TL;DR dataset with Llama-2-7B}
    \begin{tabular}{p{0.16\linewidth}p{0.83\linewidth}}
         \hline
         \textbf{Prompt} & \texttt{Am I being selfish for not wanting to leave my family to live near partner's? I've been in a long distance relationship for the last 5 years. We live over 250 miles away from each other, and have done since the beginning, apart from a few years when I was at uni, I lived a lot closer to my SO. After uni I moved back to my home town, riddled with guilt that I barely saw my family in 3 years, because I spent almost every weekend at my partners (which I feel extremely guilty about). We have always been ok with the distance, there are obviously times when it's harder than others. We're now seeing less and less of each other and I'm really starting to struggle, I'm lucky if I get a whole weekend with him every month. He works 6 days a week, so him staying with me is almost out of the question, and I can't afford any more than I already pay on travel. We have discussed getting a place together, but the issue of where to live becomes an issue. I have looked at properties in his area, and for a start I wouldn't be able to afford rent, never mind the increased cost of living.  I told him tonight that I don't think I could live far from my family. My parents are older than his, and I worry about them not having any family near them as they get older. He, on the other hand has an extremely large extended family, all of which live within a 5 mile radius. Am I being selfish for now wanting to move near them? When I told my SO how I felt, his response was well, we have a problem\". He did however say that he would move up to me, but he still has courses he needs to complete with work. He told me that I'm acting like a child and didn't understand the issue of me moving away because I would still see my parents every fortnight. Surely if travelling weren't an issue I would see my SO more often than that?! I'm feeling so conflicted, and feel selfish for not wanting to leave my parents. I love my SO, and I want to be with him, but we have hit this major fork in the road and I don't know what to do.} \\
         \hline
    \end{tabular}
\end{table}
\begin{table}[H]
    \centering
    \begin{tabular}{p{0.16\linewidth}p{0.83\linewidth}}
         \hline
         \textbf{PPO+XRLHF} & \texttt{I'm in a 5-year long-distance relationship and struggling with the idea of moving near my partner’s family. I feel guilty for not seeing my aging parents much during uni and want to stay close to them. My partner can’t travel often due to work, and I can’t afford to move to his area. He thinks I’m overreacting, but I’m torn between love and family. Am I selfish for wanting to stay near home? We’re at a crossroads and don’t know what to do.}\\
         \vspace{0.2cm}&\vspace{0.2cm}\\
         \textbf{ReMax+XRLHF} & \texttt{After 5 years in a long-distance relationship, I’m torn between moving closer to my partner or staying near my aging parents. I can’t afford to live in his area, and he can’t visit often due to work. I feel guilty about time lost with family during uni and worry they’ll need me nearby. He says we have a problem and called me childish for not wanting to move. I love him, but I’m conflicted—am I being selfish for prioritizing my family?}\\
        \hline 
    \end{tabular}
\end{table}

\subsection{Ablation study} \label{subsec: ablation study}
Recall from Sec. \ref{sec: unlearning} that we want to improve the responses to the prompts in $\bar{\mathcal{D}}_{\text{u}}$ and meanwhile do not significantly degrade the responses to the prompts in $\bar{\mathcal{D}}\setminus\bar{\mathcal{D}}_{\text{u}}$. Note that according to the no free lunch theorem \cite{adam2019no}, it is impossible to improve the responses to the prompts in $\bar{\mathcal{D}}_{\text{u}}$ without degrading the responses to the prompts in $\bar{\mathcal{D}}\setminus\bar{\mathcal{D}}_{\text{u}}$ at all. The best we can do is to avoid significant degradation on the prompts in $\bar{\mathcal{D}}\setminus\bar{\mathcal{D}}_{\text{u}}$. For this purpose, we introduce the KL-divergence term in \eqref{eq: policy unlearning} where we restricts the divergence from $\pi_0$ for the prompts in $\bar{\mathcal{D}}\setminus\bar{\mathcal{D}}_{\text{u}}$ to reduce degradation for the prompts in $\bar{\mathcal{D}}\setminus\bar{\mathcal{D}}_{\text{u}}$. To show the effectiveness of this KL divergence, we conduct an ablation study where we do not include this KL divergence in \eqref{eq: policy unlearning}.
\begin{table}[H]
\centering
\caption{Win rates of PPO+XRLHF (ReMax+XRLHF) over PPO (ReMax) on $\bar{\mathcal{D}}_{\text{u}}$ without the KL divergence.}
\label{tab: comparison on unsatisfactory no KL}
\begin{tabular}{ccccc}
\toprule
\multirow{2}{*}{Model \& Dataset} & \multicolumn{2}{c}{PPO+XRLHF vs PPO} & \multicolumn{2}{c}{ReMax+XRLHF vs ReMax} \\
& Reward (\%) & GPT-4 (\%) & Reward (\%) & GPT-4 (\%) \\
\midrule
opt-1.3B \& full-hh-rlhf & 88.2 & 84.1 & 82.8 & 89.5 \\
pythia-2.8B \& full-hh-rlhf & 87.3 & 89.0 & 90.7 & 87.2 \\
pythia-2.8B \& TL;DR & 92.4 & 89.3 & 88.1 & 87.5 \\
Llama-2-7B \& TL;DR & 89.2 & 92.2 & 86.6 & 89.4 \\
\bottomrule
\end{tabular}
\end{table}
Table \ref{tab: comparison on unsatisfactory no KL} shows that both PPO+XRLHF and ReMax+XRLHF (without the KL divergence) can significantly outperform their base RLHF algorithms on $\bar{\mathcal{D}}_{\text{u}}$. The improvement margin is larger than that in Table \ref{tab: comparison on unsatisfactory} because in this case, the policy finetuning phase does not consider the prompts in $\bar{\mathcal{D}}\setminus\bar{\mathcal{D}}_{\text{u}}$ and only aims to improve responses to the prompts in $\bar{\mathcal{D}}_{\text{u}}$
\begin{table}[H]
\centering
\caption{Win rates of PPO+XRLHF (ReMax+XRLHF) over PPO (ReMax) on $\bar{\mathcal{D}}$ without the KL divergence.}
\label{tab: comparison on validation no KL}
\begin{tabular}{ccccc}
\toprule
\multirow{2}{*}{Model \& Dataset} & \multicolumn{2}{c}{PPO+XRLHF vs PPO} & \multicolumn{2}{c}{ReMax+XRLHF vs ReMax} \\
& Reward (\%) & GPT-4 (\%) & Reward (\%) & GPT-4 (\%) \\
\midrule
opt-1.3B \& full-hh-rlhf & 43.7 & 39.2 & 49.2 & 36.6 \\
pythia-2.8B \& full-hh-rlhf & 50.8 & 45.5 & 32.0 & 42.9 \\
pythia-2.8B \& TL;DR & 38.2 & 32.4 & 35.2 & 32.8 \\
Llama-2-7B \& TL;DR & 38.5 & 40.9 & 33.4 & 37.9 \\
\bottomrule
\end{tabular}
\end{table}
Table \ref{tab: comparison on validation no KL} shows that both PPO+XRLHF and ReMax+XRLHF (without the KL divergence) become worse than their base RLHF algorithms on $\bar{\mathcal{D}}$. The reason is that, without the KL divergence regularization, PPO+XRLHF and ReMax+XRLHF easily overfit on $\bar{\mathcal{D}}_{\text{u}}$. Although PPO+XRLHF and ReMax+XRLHF significantly improve on $\bar{\mathcal{D}}$, its generalization becomes worse.

\section{Limitations}\label{appendixsec: limitations}
While XRLHF can improve the response generated by the language model. A more useful scenario is that it can improve response during test time. Specifically, the user provides feedback that the response is unsatisfactory and then the language model regenerates a better response. In this case, it requires our algorithm to quickly retrieve the training data that lead to this unsatisfactory response and then unlearn to regenerate. We will explore methods that are suitable to the test-time scenario in future works.

\section{Societal impact}\label{sec: societal impact}
Given the successful deployment of large language models (LLMs) in various human-related real-world applications, it is crucial to ensure that the responses of a tuned LLM to prompts are aligned with human or societal values and preferences, which can potentially yield direct social impacts. In our case, a malicious entity may unlearn useful data and retain harmful data to poison the LM.


\newpage
\section*{NeurIPS Paper Checklist}

\begin{enumerate}

\item {\bf Claims}
    \item[] Question: Do the main claims made in the abstract and introduction accurately reflect the paper's contributions and scope?
    \item[] Answer: \answerYes{} 
    \item[] Justification: We include a contribution statement that accurately reflects the paper's contributions and scope.

\item {\bf Limitations}
    \item[] Question: Does the paper discuss the limitations of the work performed by the authors?
    \item[] Answer: \answerYes{} 
    \item[] Justification: We include limitations in Appendix \ref{appendixsec: limitations}.

\item {\bf Theory assumptions and proofs}
    \item[] Question: For each theoretical result, does the paper provide the full set of assumptions and a complete (and correct) proof?
    \item[] Answer: \answerYes{} 
    \item[] Justification: We provide proof in Appendix \ref{sec: proof of computational complexity of explanation}.

    \item {\bf Experimental result reproducibility}
    \item[] Question: Does the paper fully disclose all the information needed to reproduce the main experimental results of the paper to the extent that it affects the main claims and/or conclusions of the paper (regardless of whether the code and data are provided or not)?
    \item[] Answer: \answerYes{} 
    \item[] Justification: We provide enough details for reproduction.

\item {\bf Open access to data and code}
    \item[] Question: Does the paper provide open access to the data and code, with sufficient instructions to faithfully reproduce the main experimental results, as described in supplemental material?
    \item[] Answer: \answerYes{} 
    \item[] Justification: We include code in supplementary materials.

\item {\bf Experimental setting/details}
    \item[] Question: Does the paper specify all the training and test details (e.g., data splits, hyperparameters, how they were chosen, type of optimizer, etc.) necessary to understand the results?
    \item[] Answer: \answerYes{} 
    \item[] Justification: We include experiment details in Appendix \ref{appendixsec: experiment details}.

\item {\bf Experiment statistical significance}
    \item[] Question: Does the paper report error bars suitably and correctly defined or other appropriate information about the statistical significance of the experiments?
    \item[] Answer: \answerYes{} 
    \item[] Justification: We provide experiment results that support our claims.
    \item[] Guidelines:
    \begin{itemize}
        \item The answer NA means that the paper does not include experiments.
        \item The authors should answer "Yes" if the results are accompanied by error bars, confidence intervals, or statistical significance tests, at least for the experiments that support the main claims of the paper.
        \item The factors of variability that the error bars are capturing should be clearly stated (for example, train/test set, initialization, random drawing of some parameter, or overall run with given experimental conditions).
        \item The method for calculating the error bars should be explained (closed form formula, call to a library function, bootstrap, etc.)
        \item The assumptions made should be given (e.g., Normally distributed errors).
        \item It should be clear whether the error bar is the standard deviation or the standard error of the mean.
        \item It is OK to report 1-sigma error bars, but one should state it. The authors should preferably report a 2-sigma error bar than state that they have a 96\% CI, if the hypothesis of Normality of errors is not verified.
        \item For asymmetric distributions, the authors should be careful not to show in tables or figures symmetric error bars that would yield results that are out of range (e.g. negative error rates).
        \item If error bars are reported in tables or plots, The authors should explain in the text how they were calculated and reference the corresponding figures or tables in the text.
    \end{itemize}

\item {\bf Experiments compute resources}
    \item[] Question: For each experiment, does the paper provide sufficient information on the computer resources (type of compute workers, memory, time of execution) needed to reproduce the experiments?
    \item[] Answer: \answerYes{} 
    \item[] Justification: We indicate the GPUs we use in Appendix \ref{appendixsec: experiment details}.
    
\item {\bf Code of ethics}
    \item[] Question: Does the research conducted in the paper conform, in every respect, with the NeurIPS Code of Ethics \url{https://neurips.cc/public/EthicsGuidelines}?
    \item[] Answer: \answerYes{} 
    \item[] Justification: We follow the Code of Ethics.

\item {\bf Broader impacts}
    \item[] Question: Does the paper discuss both potential positive societal impacts and negative societal impacts of the work performed?
    \item[] Answer: \answerNA{} 
    \item[] Justification: The societal impact is discussed in Appendix \ref{sec: societal impact}.
    
\item {\bf Safeguards}
    \item[] Question: Does the paper describe safeguards that have been put in place for responsible release of data or models that have a high risk for misuse (e.g., pretrained language models, image generators, or scraped datasets)?
    \item[] Answer: \answerNA{} 
    \item[] Justification: The paper poses no such risks.

\item {\bf Licenses for existing assets}
    \item[] Question: Are the creators or original owners of assets (e.g., code, data, models), used in the paper, properly credited and are the license and terms of use explicitly mentioned and properly respected?
    \item[] Answer: \answerYes{} 
    \item[] Justification: We clearly cite the open source datasets and models we use.

\item {\bf New assets}
    \item[] Question: Are new assets introduced in the paper well documented and is the documentation provided alongside the assets?
    \item[] Answer: \answerYes{} 
    \item[] Justification: We provide documents in the supplementary materials.

\item {\bf Crowdsourcing and research with human subjects}
    \item[] Question: For crowdsourcing experiments and research with human subjects, does the paper include the full text of instructions given to participants and screenshots, if applicable, as well as details about compensation (if any)? 
    \item[] Answer: \answerNA{} 
    \item[] Justification: The paper does not involve crowdsourcing nor research with human subjects.

\item {\bf Institutional review board (IRB) approvals or equivalent for research with human subjects}
    \item[] Question: Does the paper describe potential risks incurred by study participants, whether such risks were disclosed to the subjects, and whether Institutional Review Board (IRB) approvals (or an equivalent approval/review based on the requirements of your country or institution) were obtained?
    \item[] Answer: \answerNA{} 
    \item[] Justification: The paper does not involve crowdsourcing nor research with human subjects.

\item {\bf Declaration of LLM usage}
    \item[] Question: Does the paper describe the usage of LLMs if it is an important, original, or non-standard component of the core methods in this research? Note that if the LLM is used only for writing, editing, or formatting purposes and does not impact the core methodology, scientific rigorousness, or originality of the research, declaration is not required.
    \item[] Answer: \answerNA{} 
    \item[] Justification: The core method development in this research does not involve LLMs as any important, original, or non-standard components.

\end{enumerate}

\end{document}